\documentclass[]{article}
\usepackage{proceed2e}
\usepackage{times}
\newtheorem{THEOREM}{Theorem}[section]
\newenvironment{theorem}{\begin{THEOREM} \hspace{-.85em} {\bf :} }%
                        {\end{THEOREM}}
\newtheorem{LEMMA}[THEOREM]{Lemma}
\newenvironment{lemma}{\begin{LEMMA} \hspace{-.85em} {\bf :} }%
                      {\end{LEMMA}}
\newtheorem{COROLLARY}[THEOREM]{Corollary}
\newenvironment{corollary}{\begin{COROLLARY} \hspace{-.85em} {\bf :} }%
                          {\end{COROLLARY}}
\newtheorem{PROPOSITION}[THEOREM]{Proposition}
\newenvironment{proposition}{\begin{PROPOSITION} \hspace{-.85em} {\bf :} }%
                            {\end{PROPOSITION}}
\newtheorem{DEFINITION}[THEOREM]{Definition}
\newenvironment{definition}{\begin{DEFINITION} \hspace{-.85em} {\bf :} \rm}%
                            {\end{DEFINITION}}
\newtheorem{CLAIM}[THEOREM]{Claim}
\newenvironment{claim}{\begin{CLAIM} \hspace{-.85em} {\bf :} \rm}%
                            {\end{CLAIM}}
\newtheorem{EXAMPLE}[THEOREM]{Example}
\newenvironment{example}{\begin{EXAMPLE} \hspace{-.85em} {\bf :} \rm}%
                            {\end{EXAMPLE}}
\newtheorem{REMARK}[THEOREM]{Remark}
\newenvironment{remark}{\begin{REMARK} \hspace{-.85em} {\bf :} \rm}%
                            {\end{REMARK}}
\newcommand{\thm}{\begin{theorem}}
\newcommand{\lem}{\begin{lemma}}
\newcommand{\pro}{\begin{proposition}}
\newcommand{\dfn}{\begin{definition}}
\newcommand{\rem}{\begin{remark}}
\newcommand{\xam}{\begin{example}}
\newcommand{\cor}{\begin{corollary}}
\newcommand{\prf}{\noindent{\bf Proof:} }
\newcommand{\ethm}{\end{theorem}}
\newcommand{\elem}{\end{lemma}}
\newcommand{\epro}{\end{proposition}}
\newcommand{\edfn}{\bbox\end{definition}}
\newcommand{\erem}{\bbox\end{remark}}
\newcommand{\exam}{\bbox\end{example}}
\newcommand{\ecor}{\end{corollary}}
\newcommand{\eprf}{\bbox\vspace{0.1in}}
\newcommand{\beqn}{\begin{equation}}
\newcommand{\eeqn}{\end{equation}}
\newcommand{\bbox}{\vrule height7pt width4pt depth1pt}
\newcommand{\clm}{\begin{claim}}
\newcommand{\eclm}{\end{claim}}
\newcommand{\union}{\cup}
\newcommand{\IR}{\mbox{$I\!\!R$}}

\newcommand{\IN}{\mbox{$I\!\!N$}}

\renewcommand{\phi}{\varphi}
\newcommand{\A}{{\cal A}}

\renewcommand{\>}{\rangle}

\newcommand{\ol}{\setlength{\itemsep}{0pt}\begin{enumerate}}
\newcommand{\eol}{\end{enumerate}\setlength{\itemsep}{-\parsep}}
\newcommand{\ul}{\setlength{\itemsep}{0pt}\begin{itemize}}
\newcommand{\dl}{\setlength{\itemsep}{0pt}\begin{description}}
\newcommand{\edl}{\end{description}\setlength{\itemsep}{-\parsep}}
\newcommand{\eul}{\end{itemize}\setlength{\itemsep}{-\parsep}}

\newcommand{\commentout}[1]{}

\newcommand{\bi}{\begin{itemize}}
\newcommand{\ei}{\end{itemize}}
\newcommand{\be}{\begin{enumerate}}
\newcommand{\ee}{\end{enumerate}}

\newcommand{\denselist}{\itemsep 0pt\partopsep 0pt}

\renewcommand{\dfn}{\begin{definition}}
\renewcommand{\edfn}{\end{definition}}
\renewcommand{\xam}{\begin{example}}
\renewcommand{\exam}{\end{example}}
\renewcommand{\thm}{\begin{theorem}}
\renewcommand{\ethm}{\end{theorem}}
\renewcommand{\cor}{\begin{corollary}}
\renewcommand{\ecor}{\end{corollary}}
\renewcommand{\lem}{\begin{lemma}}
\renewcommand{\elem}{\end{lemma}}
\renewcommand{\pro}{\begin{proposition}}
\renewcommand{\epro}{\end{proposition}}
\newcommand{\RMAX}{\mbox{{\sc Rmax}}}
\newcommand{\URMAX}{\mbox{{\sc URmax}}}
\newcommand{\citeyear}{\cite}

\newcommand{\Opt}{{\rm Opt}}
\newlength{\citeskipup}
\newlength{\citeskipdown}

\renewcommand{\A}{A}
\newcommand{\shortv}{\commentout}
\newcommand{\fullv}[1]{#1}
\title{MDPs with Unawareness}
\author{Joseph Y. Halpern \ \ \ \  Nan Rong\ \ \ \ Ashutosh Saxena\\
Computer Science Department\\  
Cornell University\\ 
Ithaca, NY 14853\\
\mbox{$\{$halpern $\mid$ rongnan $\mid$ asaxena$\}$@cs.cornell.edu} }
\begin{document}
%nan20: changed to '\makeanontitle' according to NIPS format requirement
%nan21: recovered to UAI style
\maketitle
%\makeanontitle
%nan13: renewed abstract 
\begin{abstract}
Markov decision processes (MDPs) are widely used for modeling decision-making 
problems in robotics, automated control, 
and economics. Traditional MDPs assume that the decision maker (DM) knows 
all states and actions. However, this may not be true in many situations 
of interest. We define a new framework, \emph{MDPs with unawareness} (MDPUs) 
to deal with the possibilities that a DM may not be aware of all possible actions. We
provide a complete characterization of when a DM can learn to play near-optimally
in an MDPU, and give an algorithm that learns to play
near-optimally when it is possible to do so, as efficiently as possible.
%joe11
%Our results show that there will be time when an agent cannot hope to learn
%near-optimally. On the other hand, there are also class of MDPUs where 
In particular, we characterize when
a near-optimal solution can be found in polynomial time. 
\end{abstract}

%nan13: commented for UAI
%\category{I.2.8}{Artificial Intelligence}{Problem Solving, Control
%Methods, and Search}[plan execution, formation, and generation] 
%\category{F.2.0}{Theory of Computation}{Nonnumerical Algorithms and
%Problems} 
%
%\terms{Theory, Algorithms}
%
%\keywords{Markov decision process, unawareness, complexity,
%reinforcement learning, efficiency of learning, decision making.} 

%nan13: changed all sections and subsection names into capital letters according to UAI
%nan20: changed all sections and subsection names into small letters according to NIPS 
%nan21: recovered to UAI style
%\section{Introduction}\label{sec:intro}
\section{INTRODUCTION}\label{sec:intro}
%joe1*: one of the key contributions is MDPs with unawareness.  Let's
%start by stressing that
%nan10: references added
%joe9: this can be shortened for the abstract, to save space
%joe11: again, saving space
%Markov Decision Processes (MDPs)  \cite{Bellman57,Howard60,LDK95} have
%joe10
%Markov Decision Processes (MDPs)  \shortv{\cite{Bellman57}}
Markov decision processes (MDPs)  \shortv{\cite{Bellman57}}
\fullv{\cite{Bellman57,Howard60,LDK95}} have 
been used in a wide variety of 
%Markov Decision Processes (MDPs) have been used in a wide variety of
settings to model decision making.  The description of an MDP includes a
set $S$ of possible states and a set $A$ of actions.  Unfortunately, in
many decision problems of interest, the decision maker (DM) does not
know the state space, and is unaware of possible actions she can
perform.   For example, someone buying insurance may not be aware of all
possible contingencies; someone playing a video game may not be aware of
all the actions she is allowed to perform nor of all states in the
game.  
%joe3*: cut this for now; added more discussion below
%(Note that this is different from assigning a low probability to
%these states.  We are thinking more of the situation where the game
%player can be genuinely surprised by a state that arises in the course
%of play---the possibility that such a state could arise never occurred
%to her.)  

%joe4: changed ``agent'' to DM globally
%The fact that the agent may not be aware of all states does not cause 
The fact that the DM may not be aware of all states does not cause 
major problems. 
%nan3: checked: suppose we know all actions but not all states, we can 
%	always start with state number N = |S_0|, then repeated the algorithm
% with incremental values of N. Sooner or later, we will reach $N$, after
% which the algorithm continuous to achieve optimal reward.
%[[CHECK]].  
If an action leads to a new state and the set of possible
actions is known, we can use standard techniques (discussed below) to
%nan10: reference added 
%joe9: this reference isn't necessary, since we discuss the standard
%techniques below
%decide what to do next \cite{BT02}.  The more interesting issue comes in
decide what to do next.  The more interesting issue comes in
dealing  
%decide what to do next.  The more interesting issue comes in dealing
with actions that the DM may not be aware of.  If the DM is not aware of
her lack of awareness then it is clear how to proceed---we can simply
ignore these actions; they are not on the DM's radar screen.  We are
interested in a situation where the DM realizes that there are 
%joe4
%actions (and states) that he is not aware of, and thus will want to
actions (and states) that she is not aware of, and thus will want to
explore the MDP.  We model this by using a special \emph{explore}
action.  As a result of playing this action, the DM might become aware
of more actions, whose effect she can then try to understand.

%%nan13: commented out for now, to cut size and be more focused on the
%%actual model and the results we got 
%nan20: changed the next paragraph from commented to fullv
%joe14*: I think what has become clear in the last round of comments is
%that we need to be very clear about how widely applicable the model is;
%this means that we need to have *more* discussion about it, not less!
%\fullv{
%joe3*: 
We have been deliberately vague about what it means for a DM to be
``unaware'' on an action.  We have in mind a setting where there is a
%joe7: changed \A' to \A^* here and everywhere else
(possibly large) space $\A^*$ of \emph{potential actions}.  For example,
%joe4: included material from the paragraph you wrote below
in a video game, the space of potential actions may consist of all
possible inputs from all input devices combined
(e.g., all combinations of mouse movements, presses of keys on the keyboard, 
and eye movements in front of the webcam); if a DM
is trying to prove a theorem, at least in principle, all possible proof
techniques can be described in English, so the space of potential actions
%joe7
%is a subset of the set of English texts.  Of course, a
%DM's conception of potential acts may not correspond to $\A'$.
can be viewed as a subset of the set of English texts.  
%joe14*: added next sentence
The space $\A$ of \emph{actual actions} is the (typically small) subset
of $\A^*$ that are the ``useful actions''.  For example, in a video
%nan21: extra ')' removed
%joe15
%game these would be the combinations of arrow presses (and perhaps head
game, these would be the combinations of arrow presses (and perhaps head
%game) these would be the combinations of arrow presses (and perhaps head
movements) that have an appreciable effect on the game.
Of course, $\A^*$
%nan9
may not describe how the DM conceives of the potential acts.
%may not describe how the DM's conceives of the potential acts.
For example, a first-time video-game player may 
consider the action space to include only presses of the arrow keys, and be
completely unaware that eye movement is an action.  Similarly, a
mathematician trying to find a proof probably does not think of herself as
searching in a space of English texts; she is more likely to be
exploring the space of ``proof techniques''.   A sophisticated
mathematician or video game player will have a better understanding of
the space that she views herself as exploring.
Moreover, the space of potential
actions may change over time, as the DM becomes more sophisticated.
Thus, we do not explicitly describe
$\A^*$ in our formal model, and abstract the process of exploration by
just having an \emph{explore} action.  
%joe8*: cut for abstract.  We should add this issue to the conclusion too.
\fullv{(It actually may make sense to have
several different \emph{explore} actions; we defer discussion of this
point to Section~\ref{sec:conc}.)}

%joe3: 
%joe11: reinstated some of this 
%As the examples above suggest,
This type of exploration occurs all the time.  
In video games, first-time
players often try to learn the game by exploring the space of moves,
without reading the instructions (and thus, without being aware of all
the moves they can make).  Indeed, in many games, there may not be
instructions at all (even though players can often learn what moves are
available by checking various sites on the web).  
%joe11
Mathematicians trying to generate new approaches to proving a theorem
can be viewed as exploring the space of proof techniques.
%joe3
%Another example that
%academics are well aware of comes in trying to prove a theorem.  Here
%the ``moves'' are proof techniques.  Someone trying to prove a theorem
%may not be aware of how to prove it.  In research, proving a theorem
%may well involve coming up with a new proof technique (as well as trying
%out known proof techniques---these can be thought of as the moves that
%the DM is aware of).
%joe3
%In robotics, the space of potential moves for a robot with ten degrees
%of flexiblity is huge; again, exploration is almost surely necessary to
%understand the actions needed to move from one point to another.  
%%nan4*: added the Mars rover's example here
%For example, in the Mars rover project, the rover has to explore over
%a huge configuration space for new
%%joe4
%%useful actions that can be perform in an unfamiliar environment.
%useful actions that can be performed in an unfamiliar environment.
%Ash
%joe11
More practically, in robotics, if we take an action to be a ``useful''
sequence of basic moves, 
%In robotics, the space of potential moves for a robot is often huge.
the space of potential actions is often huge.
For instance, most humanoid robots (such as Honda Asimo robot \cite{SWAMHF02})
have more than 20 degrees of freedom; in such a large space, while
robot designers can hand-program a few basic actions (e.g.,
as walking on a level surface), it is practically impossible to do so
for other 
general scenarios (e.g., walking on uneven rocks).
%joe5*: we need to connect the huge exploration space to lack of
%awareness (OK)
%nan10: changed 'parameters' to 'robot configuration' to be more
%specific (suggested by Ash) 
%joe11
%We can think of the space of possible robot configurations as the
%%We can think of the space of possible settings of the parameters as the
%set $\A^*$ of potential actions.  But again, not all of these are
%necessarily actions that can be performed, and it may be conceptually
%useful to think of the designer as not being aware of the actions that
%can be performed.  
Conceptually, it is 
useful to think of the designer as not being aware of the actions that
can be performed.  
%In these cases, exploration is almost surely necessary to
%nan10: changed to 'perform the new tasks' to be more specific
%(suggested by Ash) 
%Therefore, 
Exploration is almost surely necessary to
discover new actions necessary to enable the robot to perform the new tasks.
%Exploration is almost surely necessary to
%discover new actions necessary to enable the robot to do so, 
%joe5
%so our results apply in this setting too.
%}

%joe3
Given the prevalence of MPDUs---\emph{MDPs with unawareness}, 
%joe9: shortened
%it becomes important to consider 
%whether it is possible to learn to play well in them, and, if it is, how
%that can be done.
the problem of learning to play well in an MDPU becomes of interest.
There has already been a great
%nan13: changed the format of citeyear to UAI style, and here after.
deal of work on learning to play optimally in an MDP.  
%joe11*: Nan, please undo this, and use \citeyear; I don't think there's
%a required UAI citation style, and, in any case, this shouldn't be done
%by hand; we can just change the citation style
%(Kearns and Singh, 2002) 
Kearns and Singh \citeyear{KS02}
gave an algorithm called $E^3$ that converges to
%deal of work on learning to play optimally in an MDP.  Kearns and Singh's
%\citeyear{KS02} gave an algorithm called $E^3$ that converges to
near-optimal play in polynomial time.  
%joe11: again, please undo all these changes
%(Brafman and Tennenholtz, 2002); I wish you ahd marked them!
Brafman and Tennenholtz \citeyear{BT02}
%joe11: use R-max consistently
%later gave an elegant algorithm they called R-MAX that
later gave an elegant algorithm they called $\RMAX$ that
converges to near-optimal play in polynomial time not just in MDPs, but
in a number of adversarial settings.  Can we learn to play
near-optimally in an MDPU?  (By ``near-optimal play'', we mean near-optimal
play in the actual MDP.)  In the earlier work, near-optimal play
involved learning the effects of actions (that is, the transition
probabilities induced by the action).  In our setting, the DM still has
%joe7
%to learn the transition probabilities, but it also has to learn what
%joe9
%to learn the transition probabilities, but she also has to learn what
to learn the transition probabilities, but also has to learn what
actions are available.  

Perhaps not surprisingly, we show that how effectively the DM can learn
optimal play in an MDPU depends on the probability of discovering new
actions. 
%joe7
%If it is too low, then we can never learn to play near-optimally.  
For example, if it is too low, then we can never learn to play
near-optimally.   
If it is a little higher, then the DM can learn to play near-optimally, but it
may take exponential time.  
%joe7
If it is sufficiently high, then the DM can learn to play near-optimally in
polynomial time.
%joe7*: rewrote story here
%Finally, if, for example, there is a 
%constant probability of discovering a new action, then we can use a
%modification of the R-MAX algorithm (called UR-max) 
%to show that the DM can learn near-optimal play in polynomial time.  
We give an expression whose value, under minimal assumptions, completely
characterizes when the DM can learn to play optimally, and how long it
will take.  Moreover, we show that a modification of the $\RMAX$ algorithm
(that we call $\URMAX$) can learn to play near-optimally if it is possible to
do so.
%nan3: corresponding results added :)
%[[NAN, I REALIZE YOU HAVEN'T PROVED ALL THESE RESULTS YET.]]  
%joe7: cut this from here; let's leave it to the discussion section
%In the first two cases, the
%DM  will typically be better off not bothering to try play optimally.
%(In the context of doing research, this amounts to not trying to prove
%that P $\ne$ NP because it is too hard to find the proof.)

%joe6*: This paragraph will need to be rewritten if you agree with my
%claims.  
There is a subtlety here.  Not only might the DM not be aware of what
%joe4
%actions can be performed in a given state, he may be unaware of
actions can be performed in a given state, she may be unaware of
\emph{how many} actions can be performed.  Thus, for example, in a state
%joe4: changed he to she throughout
%where he has discovered five actions, he may not know whether he has
where she has discovered five actions, she may not know whether she has
discovered all the actions (in which case she should not explore further)
or there are more actions to be found (in which case she should).  
%joe7: 
%The exponential-time lower bound and impossibility results hold even if the
%nan10: rewrote to make it easier for an unfamiliar reader (suggested by Ash)
In cases where the DM knows that there is only one action to be discovered, 
and what its payoff is, it is still possible that the DM never learns to play
optimally. Our impossibility results and lower bound hold even in this case.
(For example, if the action to be discovered is a proof
%Our impossibility results and lower bound hold even if the 
%DM knows that there is only one action to be discovered, and what
%its payoff is.  (For example, if the action to be discovered is a proof
that P $\ne$ NP, the DM may know that the action has a high payoff;
she just does not know what that action is.)  
%nan3: added the discussion on the bordering case 
%joe4*: wrong story
%So does the polynomial upper bound. 
%[On the other hand, if the probability of discovering a new action is
%constant, then near-optimal play can be learned even if the number
%of actions to be discovered is unknown.]
On the other hand, $\URMAX$ works even if the DM does not know how
%joe11
%many actions to be discovered is unknown.
many actions there are to be discovered.

%nan3: discussion of the exploration space
%joe4: this was out of place and somewhat redundant. I moved some of
%this material back to where I discussed \A'
\commentout{
Note that we are abstracting away the exploration space for the unaware
actions by the special \textit{explore} 
action. In the video game example, the actual exploration space includes all possible inputs from 
all input devices combined (e.g. the movement of mouse, a pressed key in the keyboard,
or even an eye movement in front of the webcam). However, a naive first-time player may 
consider a very different space which only includes the arrow keys. In this case, the eye 
movement is obviously unaware to him. If we think of proofs, 
one `possible' exploration space is all the English texts. This covers everything, but is clearly 
not how people think of the space. Thus, how people conceptualize the space is often critical, and 
may change over time. As an indirect result, the model we present here applies equally well  
to MDPs with a huge set of actions (all aware to the DM) but only a few of them are good. For example, suppose Google has 
a huge set of candidate advertisements where only a few are good. If we consider the good 
ones as actions, an undiscovered good ads would be an unaware action (although in this case Google
is actually aware of the ads, but not of its quality). Now consider a
robotic car trying to move  
over a narrow pass. This again can be considered as a MDPU problem where the unaware actions are 
the undiscovered successful movements, and the exploration space includes all the possible movements.
}
%joe4: \end{commentout}

%joe11: shortened
%Although we are not aware of any work on awareness in the context of
%MDPs, 
There has been a great deal of recent work on awareness in the
game theory literature (see, for example,
%nan10: added references for works on MDPs with large number of actions
%(suggested by Ash) 
%\cite{Feinberg04,Feinberg09,HMS07,HR06}).  None of these papers,
%joe11: shrunk to save space
%\cite{Feinberg04,Feinberg09,HMS07,HR06}).  There has also been works
%on MDPs with a large number of actions (see, for example 
\fullv{\cite{Feinberg04,Feinberg09,HR06,HMS07}).}
\shortv{\cite{Feinberg04,HR06,HMS07}).}  There has also been work
on MDPs with a large action space (see, for example 
%joe9: alphabetized
%\cite{NMKD98,DKG98}). 
\cite{DKG98,NMKD98}),
%joe13
and on finding new actions once exploration is initiated 
\cite{AN05ICML}.
%end nan10
None of these papers,
%joe11
%however, consider the problem of learning in the presence of lack of
however, considers the problem of learning in the presence of lack of
%joe12
%awareness; we believe that this paper is the first to do so.  
%joe16
%awareness; we believe that we are the first to do so.  
\fullv{awareness; we believe that we are the first to do so.}  
\shortv{awareness.}
%joe11: to save space; this can go in the conclusion
%Although
%we are starting here with single-person decision problems, we hope to
%extend these results to multi-player games.

%joe7*: this still needs to be done. 
%nan9: completed
%joe8: left to full paper
%joe15
%\fullv{
The rest of the paper is organized as follows. 
%nan22: added description for section 2
In Section~\ref{sec:pre}, we review the work on learning to play 
optimally in MDPs.
In Section~\ref{sec:model}, we describe our model of MDPUs.  We give our
impossibility results and lower bounds in Section~\ref{sec:lb}.  In 
Section~\ref{sec:urmax}, we present a general learning algorithm
(adapted from R-MAX) 
%joe8: just added comma
%for MDPU problems and give upper bounds. We conclude in
for MDPU problems, and give upper bounds. We conclude in
Section~\ref{sec:conc}. 
%joe15: the full paper can be posted on CoRR and cited here.
Missing proofs can be found in the full paper.  %@@
%comparisons

%nan13: new section added, to clear up the structure of the paper
% Brief explanation of MDP added, and 'Review of R-max' is shortened and relocated here.
% (so that Section 5 is now completely focused on UR-max).
%\section{Preliminaries}\label{sec:pre}
\section{PRELIMINARIES}\label{sec:pre}
%joe11: shortened
%In this section, we first review MDP. Then we present an existing
%near-optimal polynomial-time reinforcement learning algorithm for MDPs:
%R-max, of which we will be  
%extending later (in Section \ref{sec:urmax}) to obtain near-optimal
%learning for MDPUs. 
%joe14
\fullv{In this section, we review the work on learning to play optimally in
%nan20: added reference here
%MDPs and, specifically, Brafman and Tennenholt's R-MAX algorithm.  
MDPs and, specifically, Brafman and Tennenholt's R-MAX algorithm
\cite{BT02}.} 

%joe11: to save space
%\subsection{MDP}
\paragraph{MDPs:}
An MDP is a tuple $M= (S, A, P, R)$, where $S$ is a finite set of states;
$A$ is a finite set of actions; $P: (S \times S \times A) \rightarrow [0,1]$
is the transition probability function,
where $P(s,s',a)$ gives the transition probability from state $s$ to state $s'$ with action $a$; 
and $R: (S \times S \times A) \rightarrow \IR^+$ is the reward function, where
$R(s,s',a)$ gives the reward for playing action $a$ at state $s$ and transiting to state $s'$. 
%joe14
%Note that $P$ as a probability function, we have $\sum_{s' \in S}
Since $P$ is a probability function, we have $\sum_{s' \in S}
P(s,s',a) = 1$ for all $s\in S$ and $a\in A$. 
%joe14: Removed paragraph break to save space
%
A \emph{policy} in an MDP $(S,A, P,R)$ is a
function from histories to actions in $A$.
%\footnote{Although the optimal policy in an MDP is history independent,
%and just depends on the state, to find an optimal policy, a DM will need
%to make decisions based on history.}
Given an MDP $M = (S,A,P,R)$, let $U_M(s,\pi,T)$ denote the expected
$T$-step undiscounted average reward of policy $\pi$ started in state
$s$---that is, the expected total reward of running $\pi$ for $T$ steps,
divided by $T$.  Let $U_M(s,\pi) = \lim_{T \rightarrow \infty}
U_M(s,\pi,T)$, and let $U_M(\pi) = \min_{s \in S} U_M(s,\pi)$. 
% (We discuss the reason for the choice of ``min'' below.)   

%joe11
\paragraph{The mixing time:}
For a policy $\pi$ such that $U_M(\pi) = \alpha$, it may take a long
time for $\pi$ to get an expected payoff of $\alpha$.  For
example, if getting a high reward involves reaching a particular state
$s^*$, and the probability of reaching $s^*$ from some state $s$ is low,
then the time to get the high reward will be high.  To deal with this,
%joe11
%Kearns and Singh  argue that the running time of a learning algorithm
Kearns and Singh \citeyear{KS02} argue that the running time of a
learning algorithm 
should be compared to the time that an algorithm with full
information takes to get a comparable reward.  
Define the \emph{$\epsilon$-return mixing time of policy $\pi$} to be the
smallest value of $T$ such that $\pi$ guarantees an expected payoff of
at least $U(\pi) - \epsilon$; that is, it is the least $T$ such that
$U(s,\pi,t) \ge U(\pi) - \epsilon$ for all states $s$ and times $t \ge
T$.   Let $\Pi(\epsilon,T)$ consist of all policies whose
$\epsilon$-mixing time is at most $T$.  
%Let $\Opt(M) = \max_{\pi} U_M(\pi)$, and let 
Let $\Opt(M, \epsilon,T) = \max_{\pi \in \Pi(\epsilon,T)} U_M(\pi)$.  
%joe11
%\end

%joe11
%\subsection{A NEAR-OPTIMAL ALGORITHM FOR MDP: R-MAX}
\paragraph{{\bf $\RMAX$:}}
%nan20: added reference here according to the reviews 
%joe14: to be honest, I think this is overkill, but it can't hurt.
%Shortened for the abstract in any case
%$\RMAX$ is a model-based near-optimal polynomial-time reinforcement learning
\fullv{$\RMAX$ \cite{BT02} is a model-based near-optimal polynomial-time
reinforcement learning 
algorithm for zero-sum stochastic games (SG), which also directly applies to
standard MDPs.}
\shortv{We now briefly describe the $\RMAX$ algorithm \cite{BT02}.} 
$\RMAX$ assumes that the DM knows all the actions 
that can be played in the game, but needs to learn the
transition probabilities and reward function associated with each action.
%joe14
%It does not assume that the DM knows all states - new states might be 
It does not assume that the DM knows all states; new states might be 
discovered when playing actions at known states.  
$\RMAX$ follows an implicit ``explore or exploit'' mechanism that is biased
towards exploration. Here is the $\RMAX$ algorithm:
\begin{tabbing}
\fullv{$\RMAX(|S|,|A|,R_{\max},T,\epsilon,\delta,s_0)$:\\ \\}
\shortv{$\RMAX(|S|,|A|,R_{\max},T,\epsilon,\delta,s_0)$:\\}
%nan20: adjusted the body of the algorithm to suit wider columns in NIPS
%format 
%nan21: recovered the algorithm to UAI format
\commentout{
\quad Set $K_1(T) :=
\max((\lceil\frac{4|S|\,T\,R_{\max}}{\epsilon}\rceil)^3,\lceil -6
\ln^3(\frac{\delta }{6|S|\,|A|^2})\rceil)+1$\\
\quad Set $M' := M^0$ (the initial approximation described below)\\
\quad Compute an optimal policy $\pi'$ for $M'$\\
\quad Repeat until all action/state pairs $(s,a)$ are \emph{known}\\
\quad \quad \=Play $\pi'$ starting in state $s_0$ for $T$ steps or until some new state-action pair $(s,a)$ is known\\ 
\quad \>{\bf if} $(s,a)$ has just become \emph{known} 
				{\bf then} update $M'$ so that the  transition probabilities for $(s,a)$\\
\quad \quad \quad  are the observed frequencies, and the rewards for playing $(s,a)$ are those that have been\\
\quad \quad \quad  observed. \\
\quad \>Compute the optimal policy $\pi'$ for $M'$\\
\quad Return $\pi'$.
}
Set $K_1(T) :=
\max((\lceil\frac{4|S|\,T\,R_{\max}}{\epsilon}\rceil)^3,\lceil -6
\ln^3(\frac{\delta }{6|S|\,|A|^2})\rceil)+1$\\
Set $M' := M^0$ (the initial approximation described below)\\
Compute an optimal policy $\pi'$ for $M'$\\
Repeat until all action/state pairs $(s,a)$ are \emph{known}\\
\quad \=Play $\pi'$ 
starting in state $s_0$
for $T$ steps or until some new\\
\>\quad\=  state-action pair $(s,a)$ is known\\ 
\>{\bf if} $(s,a)$ has just become \emph{known} {\bf then} update $M'$
so that\\
\>\> the  transition probabilities for $(s,a)$ are the observed\\
\> \>frequencies and the rewards for playing $(s,a)$ are those\\
\>\>that have been observed. \\
\>Compute the optimal policy $\pi'$ for $M'$\\
Return $\pi'$.
%nan21: end recovered algorithm 
%%end nan20
\end{tabbing}
%joe12
%where $R_{\max}$ is the maximum possible reward; $\epsilon> 0$,
%joe14
%Here $R_{\max}$ is the maximum possible reward; $\epsilon> 0$,
Here $R_{\max}$ is the maximum possible reward; $\epsilon> 0$;
$0<\delta<1$; $T$ is the $\epsilon$-return mixing 
time; $K_1(T)$ represents the number of visits required to approximate a transition function; 
a state-action pair $(s,a)$ is said to be \emph{known} only if it has been
played $K_1(T)$ times. $\RMAX$ proceeds in iterations, and $M'$ is the current ``approximation'' to the true MDP.
$M'$ consists state set $S$ and a dummy state $s_d$. The transition
and reward functions in $M'$ may be different from those of the actual MDP.  
In the initial approximation $M^0$, the transition and reward functions 
are trivial: when an action $a$ is taken in any state $s$ (including the dummy state $s_d$),
with probability 1 there is a transition to $s_d$, with
reward $R_{\max}$.  %Note that all policies in the initial approximation are optimal.

%nan20: added reference here according to the reviews
%Brafman and Tennenholtz show that
Brafman and Tennenholtz \cite{BT02} show that
$\RMAX(|S|,|A|,R_{\max},$ $T,\epsilon,\delta,s_0)$ learns a policy with
expected payoff within $\epsilon$ of $\Opt(M, \epsilon,T)$
with probability greater than $1-\delta$, no matter what state $s_0$ it
starts in, in time polynomial in $|S|$, $|A|$, $T$, $1/\delta$, and
$1/\epsilon$. 
%joe14: removed paragraph break
%
What makes $\RMAX$ work is that in each iteration, it either
achieves a near-optimal reward with respect to the real model or learns
an unknown transition with high probability. Since there are only
polynomially-many $(s,a)$ pairs (in the number of states and actions) to
learn, and each transition entry requires $K_1(T)$ samples, where $K_1(T)$ is
polynomial in the number of states and actions, $1/\epsilon$,
$1/\delta$, and the $\epsilon$-return mixing time $T$,
$\RMAX$ clearly runs in  time polynomial in these parameters.
In the case that the $\epsilon$-return mixing time $T$ is not known, 
%joe11
%$\RMAX$ starts with $T=1$, then consider $T=2$, $T=3$, and so on.
$\RMAX$ starts with $T=1$, then considers $T=2$, $T=3$, and so on.
%joe11
%We expand on this point in more detail in Section \ref{sec:urmax}.
\fullv{
We expand on this point further in Section \ref{sec:urmax},
for in an MDPU we need to deal with the fact that the number of states and
actions is unknown.} 
%joe14
%\shortv{We expand on this point further in Section \ref{sec:urmax}.}

%\section{MDPs with unawareness}\label{sec:model}
\section{MDPS WITH UNAWARENESS}\label{sec:model}
%nan3: paragraph rewritten
Intuitively, an MDPU is like a
standard MDP except that the player is initially aware of only a subset
of the complete set of states and actions.  To reflect the fact that
%joe4
%new states and actions could be learned during the game, the
new states and actions may be learned during the game, the
model provides a special \emph{explore} action. 
By playing this action, the DM may become aware of actions
that she was previously unaware of. 
%joe4
%The probability distribution of how
%actions are discovered is represented by the discovery probability function.
%At any moment in game, the DM can only perform actions 
The model includes a \emph{discovery probability function} characterizing 
the likelihood that a new action will be discovered.
At any moment in game, the DM can perform only actions 
that she is currently aware of. 

%joe1: I think the model should allow unbounded states and actions; any
%results that require (un)boundedness should explicitly assume it.
%As its name suggests, the
%game has bounded number of states and actions (games with unbounded
%states and actions will be discussed in Section 4). Formally: 

%joe10
%\dfn A \emph{Markov Decision Process with Unawareness} is a tuple 
%nan13: 'a_0' and 'g_A' added into the tuple, and explanation of
%parameters adapted according to 
% the MDP definition given in the Preliminaries
\dfn An MDPU is a tuple $M= (S, A, S_0, a_0, g_A, g_0, P, D, R, R^+, R^-)$, where
\begin{itemize}
\denselist
\item $S$, the set of states in the underlying MDP;
\item $A$, the set of actions in the underlying MDP;
\item $S_0 \subseteq S$ is the set of states that the DM is initially
aware of; 
\item $a_0 \notin A$ is the \emph{explore} action;
\item $g_A: S \rightarrow 2^A$, where $g_A(s)$ is 
the set of actions that can be performed at $s$ other than
$a_0$ (\fullv{we assume that }$a_0$ can be performed in every state);
\item $g_0 : S_0 \rightarrow 2^{A}$, where $g_0(s) \subseteq g_A(s)$
is the set  of actions that the DM is aware of at 
state $s$ (\fullv{we assume that }the DM is always aware of $a_0$);
\item $P: \union_{s \in S} (\{s\} \times S \times g_A(s) \rightarrow
[0,1]$ is the transition probability function 
(as usual, we require  that $\sum_{s' \in S} P(s,s',a) = 1$ if $a \in g_A(s)$);
\item $D : \IN \times \IN \times S \rightarrow [0,1]$ is the discovery probability function.
$D(j,t,s)$ gives the probability of discovering a new action in
state $s\in S$ given that there are $j$ actions to be discovered and $a_0$ has
already been played $t-1$ times in $s$ without a new action being
discovered (see below for further discussion);
\item $R: \union_{s \in S} (\{s\} \times S \times g_A(s)) \rightarrow \IR^+$
%joe11
%is the reward function,
is the reward function;%
\footnote{We assume without loss of generality that all payoffs are
non-negative.  If not,
we can shift all rewards by a positive value so that all payoffs become
%joe11
%non-negative.}; 
non-negative.} 
\item 
$R^+: S  \rightarrow \IR^+$ and $R^-: S  \rightarrow \IR^+$
give the exploration reward for playing $a_0$ at state $s\in S$ and
discovering (resp., not discovering) a new action (see below for further
discussion). 
\end{itemize}
%joe8: we don't really need this
\fullv{
Let $M^u = (S,A,g,P,R)$ be the MDP \emph{underlying} the MDPU $M$.}
\edfn 
Given $S_0$ and $g_0$, we abuse
notation and take $A_0 = \union_{s \in s_0} g_0(s)$; that is, $A_0$ is
the set of actions that the DM is aware of.

Just like a standard MDP, an MDPU has a state space $S$, action space
$A$, transition probability function $P$, and reward function $R$.%
\footnote{It is often assumed that the same actions can be performed in
all states.  Here we allow slightly more generality by assuming that the
actions that can be performed is state-dependent, where the dependence
is given by $g$.}
Note that we do not give the transition function for
the explore action $a_0$ above; since we assume that $a_0$
does not result in a state change (although new actions might be discovered
%joe14
%when $a_0$ is played); for each state $s \in S$, we have
when $a_0$ is played), for each state $s \in S$, we have
$P(s,s,a_0) = 1$.   The new features here involve dealing with $a_0$.  
We need to quantify how hard it is to discover a new action.
Intuitively, this should in general depend on how many actions there are
to be discovered, and how long the DM has been trying to find a new
action.  For example, if the DM has in fact found all the actions,
then this probability is clearly 0.  Since the DM is not assumed to know
in general how many actions there are to be found, all we can do
is give what we view as the DM's subjective probability of finding a new
action, given that there are $j$ actions to be found.
%nan13: added to clarify the confusions of reviewers
Note that even if the DM does not know the number of actions, she can still
condition on there being $j$ actions.
In general, we
also expect this probability to depend on how long the DM has been
trying to find a new action.  This probability is captured by
$D(j,t,s)$.    

%joe11
%We assume $D(j,t,s)$ is nondecreasing as a function of $j$: with
We assume that $D(j,t,s)$ is nondecreasing as a function of $j$: with
more actions available, it is easier to find a new one.  How $D(j,t,s)$
varies with $t$ depends on the problem.  For example, if the DM is
%joe9: capitalized iPhone
%searching for the on/off button on her new iphone which is guaranteed to
searching for the on/off button on her new iPhone which is guaranteed to
be found 
in a limited surface area, then $D(j,t,s)$ should increase 
as a function of $t$. The more possibilities have been eliminated, the
more likely it is that the DM will find the button when the next
possibility is tested.  On the other hand, if the DM is searching for a
proof, then the longer she searches without finding one, the more
discouraged she will get; she will believe that it is more likely   
that no proof exists. In this case, we would expect 
$D(j,t,s)$ to decrease as a function of $t$.  Finally, if we think of
the \emph{explore} action as doing a random test in some space of
potential actions, the probability of finding a new action is a
constant, independent of $t$.  In the sequel, we assume for ease of
exposition that 
%nan3
%joe11
%the value of 
$D(j,t,s)$ is independent of $s$, 
%joe4
%and the function becomes $D(j,t)$. 
so we write $D(j,t)$ rather than $D(j,t,s)$.  

%[[NOW WE NEED A DISCUSSION OF THE REWARD FUNCTION FOR $a_0$]]
%nan3: a discussion of the reward function for $a_0$.
%joe4*: what you wrote isn't true any more.  In any case, we need more
%discussion here; see the next %joe3*.
%[[NAN, WE NEED MORE DISCUSSION HERE!  I ADDED SOME.  HOW MUCH WOULD IT
%AFFECT THINGS IF $R(s,s^+,a_0)$ and $R(s,s^-,a_0)$ DEPENDED ON THE STATE?]]
% nan5: the dependence of $R(s,s^+,a_0)$ and $R(s,s^-,a_0)$ on the
% states shouldn't affect 
% the existing results. 
%joe5: In that case, I think we should have $R^+(s,a)$ and R^-(s,a), and
%relate it to the reward function R.  What's really going on is that you
%can think of there as being two states s^+ and s^-; both of these are
%copies of s, but in s^+, the DM has discovered a new action after
%joe7: trying to consistently call the DM ``she''; Nan, can you check for
%consistency everywhere else?
%nan9: consistency checked
%%exploration, and in s^- he hasn't.  So R^+(s,a) is really R(s,s^+,a),
%exploration, and in s^- she hasn't.  So R^+(s,a) is really R(s,s^+,a),
%and R^-(s,a) is really R(s,s^-,a).  Presenting it this way will make it
%seem more in the keeping of standard MDPs, which I think is a good thing.
%nan6: changed the discussion according to new notations
%joe6
%$R(s,s^+,a_0)$ and $R(s,s^-,a_0)$ are the analogue of the reward
%function
$R^+$ and $R^-$ are the analogues of the reward function
$R$ for the \emph{explore} action $a_0$, 
%joe14
\fullv{
Although performing action $a_0$ does not change the state, we can
think of there being two copies of each state $s$, call them
$s^+$ and $s^-$, which are just like $s$ except that in $s^+$ the DM has
discovered a new action after exploration, and in $s^-$
she hasn't.  Then $R^+(s)$ and $R^-(s)$ can be thought of as 
$R(s,s^+,a_0)$ and $R(s,s^-,a_0)$, respectively.
In a chess game,}
\shortv{For example, in a chess game,} 
the \emph{explore} action corresponds to thinking.   
There is clearly a negative reward to thinking and not discovering a new
action---valuable time is lost; we capture this by $R^-(s)$.  
%joe9
%On the other hand, the player often gets a thrill if a useful action is
On the other hand, a player often gets a thrill if a useful action is
discovered;  
and this is captured by $R^+(s)$. It
%In a chess game, the \emph{explore} action corresponds to thinking.  
%There is clearly a negative reward to thinking and not discovering a new
%action---valuable time is lost; we capture this by $R^-(s)$.  
%It
%joe6: what does it follow from?
%natually follows that $R(s,s^-,a_0)\le R(s,s^+,a_0)$.
seems reasonable to require that $R^-(s)\le R^+(s)$, which we do from
%nan10: 'in' removed
here on.  
%joe12: to save space
\fullv{
Since
%here on in.  Since
the whole point of exploration is to discover a new action, the reward
for discovering one should be greater than the reward for not
discovering one.
}
When an MDPU starts, $S_0$ represents the set of states that the DM
is initially aware of, and $g_0(s)$ represents the set of actions that
she is aware of at state $s$. The DM may discover new states when trying out 
known actions, she may also discover new actions as the explore action $a_0$ is played.
At any time, the DM has a current set of states and actions that she is
%joe4
%aware of, and  she can only play actions from the set that she is
aware of;  she can play only actions from the set that she is
currently aware of.

In stating our results, we need to be clear about what the inputs to an
algorithm for near-optimal play are.
We assume that $S_0$, $g_0$, $D$, $R^+$, and $R^-$ are always part of
the input to the algorithm.  The reward function $R$ is not given, but
is part of what is learned.  (We could equally well assume that $R$ is
%joe7
%given for actions in $\A_0$ and states in $S_0$.)  Brafman and
given for the actions and states that the DM is aware of; this
assumption would have no impact on our results.)  
%joe11
%(Brafman and Tennenholtz 2002) assume that the DM is given a bound on
Brafman and Tennenholtz \citeyear{BT02} assume that the DM is given a bound on
the maximum reward, but later show that this information is not needed
to learn to play near-optimally in their setting.  
%joe7
%When a polynomial-time algorithm exists, knowing a bound on the reward
%is not needed in our setting either.  Perhaps the most interesting 
Our algorithm $\URMAX$ does not need to be given a bound on the reward 
either.  Perhaps the most interesting 
question is what the DM knows about $\A$ and $S$.  Our lower bounds and
impossibility result hold even if the DM knows $|S|$ and $|g_A(s)|$ for
all $s \in S$.  On the
other hand, $\URMAX$ requires neither $|S|$ nor $|g_A(s)|$ for $s \in S$.
That is, when 
something cannot be done, knowing the size of the set of states and
actions does not help; but when something can be done, it can be done
without knowing the size of the set of states and actions.
%joe7
%that it does not help a DM to know $|S|$, it may help the DM to know
%$|\A|$. 
%joe7
%We make it clear when proving our results whether this makes a difference.  
%[[NAN, WE NEED A DISCUSSION HERE OF WHAT THE INPUTS ARE/SHOULD BE/COULD
%BE TO AN ALGORITHM THAT'S TRYING TO LEARN OPTIMAL PLAY.]]

%joe7: moved from below, and rewrote slightly
Formally, we can view the DM's knowledge as the input to the learning
algorithm.   
An MDP $M$ is \emph{compatible with the DM's knowledge} if all the
parameters of  
of $M$ agree with the corresponding parameters that the DM knows
about. If the DM knows only $S_0$, $g_0$, $D$, $R^+$, 
%$R^-$), then every MDP ($S'$, $A'$, $P'$, $R'$) where $S_0\subseteq
%S'$ and $\A_0\subseteq A'$ is compatible. If the DM is given any extra
and $R^-$ (we assume that the DM always knows at least this), then every
%joe7
%MDP ($S'$, $A'$, $P'$, $R'$) where $S_0\subseteq 
MDP $(S', A', g', P', R')$ where $S_0\subseteq 
S'$ and $g_0(s) \subseteq A'(s)$ is compatible with the DM's
knowledge. If the DM 
%joe7
%is given any extra input, a compatible MDP must  
%agree with that extra input. For example, if $|A|$ is given as an extra 
%input, then a compatible MDP must have $|A'|=|A|$.
also knows $|S|$, then we must have $|S'| = |S|$; if the DM 
knows that $S = S_0$, then we must have $S' = S_0$.  
%joe7*: for consistency with BT; also put in the R_{max} assumption here
We use $R_{\max}$  to denote the maximum possible reward.  Thus, if the
DM knows $R_{\max}$, then in a compatible MDP, we have $R(s,s',a') \le
R_{\max}$, with equality holding for some transition.  (The DM may just
know a bound on $R_{\max}$, or not know $R_{\max}$ at all.)  If the DM
knows $R_{\max}$, we assume that $R^+(s) < R_{\max}$ for all $s \in S$
(for otherwise, the optimal policy for the MDPU becomes trivial: the DM
should just get to state $s$ and keep exploring).  
%joe7*: added the remainder of the section; it's critical, and this is
%where it belongs
Brafman and Tennenholtz essentially assume that the DM knows $|A|$,
$|S|$, and $R_{\max}$.  They say that they believe that the assumption
that the DM knows $R_{\max}$ can be removed.  It follows from our
results 
%joe14
\fullv{that, in fact, the}
\shortv{that the}
DM does not need to know any of $|A|$, $|S|$, or $R_{\max}$.

Our theorems talk about whether there is an algorithm for a DM to learn
to play near-optimally given some knowledge.  We define ``near-optimal
play'' by extending the definitions of 
%joe11
%(Brafman and  Tennenholtz, 2002)
%and (Kearns and Singh, 2002) to deal with
%joe14
\fullv{Brafman and  Tennenholtz \citeyear{BT02}
and Kearns and Singh \citeyear{KS02}}
\shortv{\cite{BT02,KS02}} 
to deal with unawareness.  
%joe11
%We review these definitions here, and explain how we
%extend them.  
%
%We will be interested in policies for MDPUs.  
In an MDPU, a policy is
again a function from histories to actions, but now the action must be
one that the DM is aware of at the last state in the history.
%joe8
%The DM \emph{can learn to play near-optimally given some
The DM \emph{can learn to play near-optimally given a state space $S_0$
and some other
knowledge} if, for all
%joe8: the policy can be state-dependent; this makes things easier (and
%is without loss of generality, since we can ``paste together'' the
%policies construted for each state.
%$\epsilon >  0$, $\delta > 0$, and $T$, there exists a 
%policy $\pi_{\epsilon,\delta,T}$ 
$\epsilon >  0$, $\delta > 0$, $T$, and $s \in S_0$, the DM can learn a 
policy $\pi_{\epsilon,\delta,T,s}$ 
such that, for all MDPs $M$ compatible with the DM's knowledge, there
exists a time $t_{M,\epsilon,\delta,T}$ such that, with probability at
%joe8
%least $1-\delta$, $U_M(s,\pi_\epsilon,t)  \ge \Opt(M,\epsilon,T) -
least $1-\delta$, $U_M(s,\pi_{\epsilon,\delta,T,s},t)  \ge \Opt(M,\epsilon,T) -
%nan10: $s$ seems to be defined above, so i removed 'for all state $s$'
\epsilon$ for all $t \ge t_{M,\epsilon,\delta,T}$.%
%\epsilon$ for all states $s$ and $t \ge t_{M,\epsilon,\delta,T}$.
%nan10: we seem to have an incomplete sentence here
%nan11: cut 
%necessary, even if the DM is aware of all states and actions.%
%joe8
\footnote{Note that we allow the policy to depend on the state.
%joe9
%However, the policy has to have an expected payoff that is close to that
However, it must have an expected payoff that is close to that
obtained by $M$ no matter what state $M$ is started in.}
The DM \emph{can learn to play near-optimally given some knowledge in
polynomial (resp., exponential) time} if, there exists a 
polynomial (resp., exponential)
function $f$ of five arguments such
that we can take $t_{M,\epsilon,\delta,T} = 
f(T,|S|,|A|,1/\epsilon,1/\delta)$.

%nan7*: change the name of section
%\section {Impossibility results and lower bounds} \label{sec:lb}
\section {IMPOSSIBILITY RESULTS AND LOWER BOUNDS} \label{sec:lb}

%Ash, added a new paragraph --- could be used somwhere else.
%joe5: moved paragraph up to the beginning of the section, and rewrote
%slightly.  
The ability to estimate in which cases the DM can learn to play 
optimally is crucial in many situations. For example, in robotics, if 
%joe14
%we know that the probability of discovering new actions is too low and 
%hence a robot would never discover the optimal policy (or it
%would require an exponential time to do so), then the designer of the robot
the probability of discovering new actions is so low that it would 
would require an exponential time to learn to play near-optimally, then
the designer of the robot 
%joe14
%would choose to have human engineers design the actions and not 
must have human engineers design the actions and not 
rely on automatic discovery.  
%joe5: added next sentence
%joe14
%Thus, we begin by trying to understand when it is feasible to learn to
We begin by trying to understand when it is feasible to learn to
play optimally, and then consider how to do so.  
%joe14: this can go in the conclusion of the full paper.  This is the
%wrong place
%joe15: We can expand the conclusions; I think that would be at least as
%useful as adding proofs of theorems
%Our methods and results
%can thus provide guiding principles for designing complex systems.
%nan21: break recovered 
%joe14: ran on paragraph

%joe3
%We show that some problems require at least infinite time to achieve the
%expected reward independent of the algorithm used. 
%joe5
%We show that, for some problems, there are no algorithms that can
We first show that, for some problems, there are no algorithms that can
%joe4
%guarantee near-optimal play in less than exponential time. This is true
guarantee near-optimal play; in other cases, there are algorithms that
will learn to play near-optimally, but will require at least exponential
time to do so.  These results hold even for 
%joe11
%problems where the DM knows there are two actions, 
problems where the DM knows that there are two actions, 
%joe4
%and already knows one of them.
already knows one of them, and knows the reward of the other.

%joe4*: put 
%nan4: define $E_t$ and $P_j$ here to reduce redundancies.
%Define $E_t$ to be the event of playing $a_0$ for $t$ times without
%discovering a new actions while there is at least one new action, and
%define $P_j$ to be the probability that there are in fact $j$ new
%actions.  
%joe7
%For these examples, the following notation is helpful.
%nan21: paragraph recovered since we have enough space in UAI
%joe14
%joe15: cut from here again
\fullv{
In the following examples and theorem, we use
$E_{t,s}$ to denote the event of playing $a_0$ $t$ times at state $s$ without
discovering a new action, conditional on there being at least one undiscovered
action.
} 
%nan7*: added to describe compatibility
%joe7
%Define an MDP to be compatible with what the DM knows if it does not
%contradict the input that is given to the DM. If the DM is given as

%nan7: removed description on P_{j,s}
%and let $P_{j,s}$ be the probability that there are in fact $j$
%undiscovered actions at state $s$.

%[[I REORDERED HERE, STARTING WITH THE IMPOSSIBILITY RESULT.  NAN, YOU
%NEED A GOOD STORY HERE.  THE ORDERING WILL DEPEND IN PART ON THE STORY.  
%RIGHT NOW YOU HAVE A BUNCH OF EXAMPLES AND THEOREMS, BUT NO REAL STORY
%TO LINK THEM TOGETHER.]]

%We now show that, for some problems, there are no algorithms that can
%guarantee near-optimal play in finite time. This is true even for
%problems where the DM knows there are two actions, and already knows one
%of them.
%joe3*: we should make this a theorem, not an example
\xam\label{case:0}
%joe3*: you need to have a R(s,s^+,a_0), R(s,s^-,a_0)
%$M = (S, A, S_0,  \A_0, P, R, D)$ is an MDPU with a single state
%$S=\left\{s_1\right\}$ and two actions $A=\left\{a_1,
%a_2\right\}$. Reward function $R(s_1,s_1,a_1)=r_1$ and
%$R(s_1,s_1,a_2)=r_2$ where $r_2-r_1=\alpha>0$. Transition function
%joe7
%Let $M = (S, A, S_0,  \A_0, P, R, D)$ be an MDPU where 
Suppose that the DM knows that $S = S_0 = \left\{s_1\right\}$, 
$g_0(s_1) = \{a_1\}$, $|\A| = 2$, 
$P(s_1,s_1,a)=1$ for all action $a \in
A$, $R(s_1,s_1,a_1)=r_1$, $R^+(s_1)=R^-(s_1)=0$, 
$D(j,t)=\frac{1}{(t+1)^2}$,  
and the reward for the optimal policy in the
%nan13: to make it clear about $r_2$
%joe11
%true MDP is $r_2$ where $r_2 > r_1$.  Since the DM
true MDP is $r_2$, where $r_2 > r_1$.  Since the DM
%true MDP is $r_2 > r_1$.  Since there the DM
knows that there is only 
%true MDP is $r_2 > r_1$.  Since there the DM knows that there is only
one state and two actions, the DM knows that in the true MDP, there is
an action $a_2$ that she is not aware of such that $R(s_1,s_1,a_2) =
r_2$.  That is, she knows everything about the true MDP but the action
$a_2$.  
%nan4: R(s,s^+,a_0) and R(s,s^-,a_0) added
%nan6: notation 
%$R(s,s^+,a_0)=R(s,s^-,a_0)=0$ 
%joe6
%$R(s,s^+,a_0)=R(s,s^-,a_0)=0$ for all $s$ 
%joe7
%$R^+(s,a_0)=R^-(s,a_0)=0$ for all $s$ 
%$R^+(s_1)=R^-(s_1)=0$, 
%and $D(j,t)=\frac{1}{(t+1)^2}$. 
%joe4
%(The MDPU in this example is the same as in example \ref{case:5}, but
%with a different $D$ function).   
%nan4: inputs added
%joe4*: slow down here!
%The following inputs are given to the DM: $S_0$, $\A_0$, $D$, and
%$|\A|$.
%joe7*
%We now show that there is no algorithm for near-optimal play for this
%MDP.  This is true even if (in addition to $S_0$, $g_0$, $D$, $R^+$,
%and $R^-$) the DM knows that $S_0 = S$, $|\A| = 2$, and how $R$ works
%nan21: recovered the full proof for UAI 
%joe14:
%\fullv{
We now show that, given this knowledge, the DM cannot learn to play
optimally.  

Clearly in the true MDP the optimal policy is to always play $a_2$.
%It is easy to check that the maximum expected return for the underlying
%%joe3
%%MDP problem is $r_2$ and the optimal strategy is to play $a_2$. However
%MDP problem is $r_2$ and the optimal policy is to always play $a_2$. 
%joe3
%we now prove that no algorithm can attain an expected return  $\geq
However, to play $a_2$, the DM must learn about $a_2$.  As we now show,
no algorithm can learn about $a_2$ with probability greater than $1/2$,
and thus no algorithm can attain an expected return  $\geq
(r_1 + r_2)/2 = r_2 - (r_2-r_1)/2$.

%joe3
%We prove that no matter what algorithm is used, it requires more than
%infinite time to discover the initially unaware action $a_2$ with
%probability greater than $1/2$. 
%nan4: moved the definition to the beginning of Section 3
%Let $E_t$ be the event of playing $a_0$ for $t$ times without discovering $a_2$. 
%joe3*: the question isn't what we know, but what the DM knows.
%discovering $a_2$. We know that there is exactly one unknown
%action. Thus, 

%joe15: moved this sentence here
%nan22: put it into shortv 
\shortv{Let $E_{t,s}$ denote the event of playing $a_0$ $t$ times at state $s$ without
discovering a new action, conditional on there being at least one
undiscovered action.}
Since there is exactly one unknown action, and the DM
knows this, we have
%[[NEED TO MAKE THIS PART OF THE ALGORITHM'S INPUT]],
%joe4: there's no need to label equations unless you actually refer to them
%\begin{eqnarray}\label{eq:0_1}	
%joe14: we could put this on one line to save space
$$\begin{array}{lll}
%joe3: no semicolon
%Pr(E_t)	&=& \prod_{t_0=1}^t(1-D(1,t));\nonumber \\
%&=& \prod_{t_0=1}^t(1-\frac{1}{(t_0+1)^2});\nonumber \\
%&=& \frac{t+2}{2(t+1)};\nonumber \\					
%joe4: using t' rather than t_0 as the index; it's more standard; made
%this change everywhere
%Pr(E_t)	&=& \prod_{t_0=1}^t(1-D(1,t))\nonumber \\
%nan6: notation 
%Pr(E_t)	&=& \prod_{t'=1}^t(1-D(1,t'))\nonumber \\
Pr(E_{t,s_1})	&=& \prod_{t'=1}^t(1-D(1,t'))\nonumber \\
&=& \prod_{t'=1}^t \left(1-\frac{1}{(t'+1)^2}\right) \nonumber \\
%joe3*: Nan, you need to explain why the product is t+2/2(t+1) 
%nan4: explanations added
%nan20: only include explanations in the full version, to save space
%&=& \frac{t+2}{2(t+1)} \ \ \ \mbox{[see below]}\nonumber \\		
&=& \frac{t+2}{2(t+1)} \nonumber \\		
&>& \frac{1}{2}.
%\end{eqnarray}			
\end{array}
$$

%nan20: move the explanation into full version, to save space
%joe14: explanation is not what we should cut!
%\fullv{
%nan4*: explanations
%joe4: slow down a little
%Since
For the third equality, note that
$1-\frac{1}{(t'+1)^2}=(1-\frac{1}{t'+1})\times(1+\frac{1}{t'+1})$;
%joe4
%thus, 
it follows that 
%joe7: undid this; it's OK as is (the slight spillover is acceptable,
%and we'll probably want the space for other things)
%nan8*: changed to array to fit into column size
%nan11: changed to plain format to save space
%joe10: it's probably better to use \fullv here; it's easier to undo
%nan12: changed to fullv/shortv
\fullv{
$$\prod_{t'=1}^t \left(1-\frac{1}{(t'+1)^2}\right)
%joe5: need larger parens
%	= (\frac{1}{2}\times\frac{3}{2})\times
%		(\frac{2}{3}\times\frac{4}{3})\times\cdots\times
%		(\frac{t}{t+1}\times\frac{t+2}{t+1}) \nonumber 
	= \left(\frac{1}{2}\times\frac{3}{2}\right)\times
		\left(\frac{2}{3}\times\frac{4}{3}\right)\times\cdots$$
		$$\times
		\left(\frac{t}{t+1}\times\frac{t+2}{t+1}\right). \nonumber $$
		}
\shortv{		
$\prod_{t'=1}^t \left(1-\frac{1}{(t'+1)^2}\right)
%joe5: need larger parens
%	= (\frac{1}{2}\times\frac{3}{2})\times
%		(\frac{2}{3}\times\frac{4}{3})\times\cdots\times
%		(\frac{t}{t+1}\times\frac{t+2}{t+1}) \nonumber 
	= \left(\frac{1}{2}\times\frac{3}{2}\right)\times
		\left(\frac{2}{3}\times\frac{4}{3}\right)\times\cdots\times
		\left(\frac{t}{t+1}\times\frac{t+2}{t+1}\right). \nonumber $
}
%joe7: cut
%	= \frac{t+2}{2(t+1)}.$$	
%joe7: added to help the reader
All terms but the first and last cancel out.  Thus, the product is 
%joe11
\fullv{$\frac{t+2}{2(t+1)}$, as desired.}
\shortv{$\frac{t+2}{2(t+1)}$.}
%end nan4
%joe14
%}%end fullv
%nan4
%This shows that $Pr(E_t)$ is always strictly greater than 1/2
%joe4
%The above inequality shows that $Pr(E_t)$ is always strictly greater
%than 1/2 
%nan6
The inequality above shows that $Pr(E_t,s_1)$ is always strictly greater
%The inequality above shows that $Pr(E_t)$ is always strictly greater
than 1/2, 
%joe3
%independent of of $t$. In other words, the player cannot guarantee to
independent of $t$. In other words, the DM cannot 
discover the better action $a_2$ with probability greater than $1/2$ no
matter how many times $a_0$ is played. 
%joe3*: this is confused
%
%Since this property is decided by the $D$ function, not by the algorithm
%used, any algorithm would require infinite time to discover $a_2$ with
%probability greater than $1/2$. If $a_2$ is not discovered, the best
%reward any algorithm can achieve is $r_1$. As a result, no algorithm may
%guarantee to attain a reward $\geq r_2-\alpha/2$ with probability
%greater than $1/2$.  
It easily follows that the expected reward of any policy is at most
%joe8
%$r_2 - \alpha/2$.  Thus, there is no algorithm that learns to play
$(r_1 + r_2)/2$.  Thus, there is no algorithm that learns to play
near-optimally. 
%joe14
\eprf
\exam
%end nan2
%}
\commentout{
%joe14
\shortv{
As we show in the full paper, 
no algorithm can learn about $a_2$ with probability greater than $1/2$,
and thus no algorithm can attain an expected return  $\geq
(r_1 + r_2)/2 = r_2 - (r_2-r_1)/2$.  Thus, there is no algorithm that
learns to play near-optimally.
\eprf
\exam
}
%joe14: \end{shortv}
}
%nan21: end recovered proof

%[[NAN: Let's talk about this theorem.  It's not clear it's worth
%proving something so general.   If you're going to prove this though, I
%think it suffices to assume that (1) $\sum_{t_1}^\infty f(t) < \infty$,
%(2) $f(1) < 1$, and (3) $f$ is differentiable.  You don't mention (2)
%and (3), but you need them for your proof.  The advantage of my
%formulation is that it makes both examples special cases -- all the more
%reason to cut the second example.  Also, you should make clear what
%input the algorithm gets.  There should be no problem knowing how many
%actions there are.]]

%nan2: added a general theorem for the impossibility result
%nan3: changed the theorem to apply to a more general class of MDPUs
%nan4*: theorem definition refined, proof restructured and rewritten
%(and redundancy removed). 
%joe4: let's bring out the story a bit better
%We give a theorem that classifies the MDPUs that require more than
%finite time to guarantee a near-optimal play no matter what algorithm is
%used. 
The problem in Example~\ref{case:0} is that the discovery probability is
so low that there is a probability bounded away from 0 that some action
will not be discovered, no matter how many times $a_0$ is played.  The
following theorem generalizes Example~\ref{case:0}, giving a sufficient
condition on the failure probability 
%joe7
(which we later show is also necessary) that captures the precise sense
in which the discovery probability is too low.
%joe7*: added; we should dicuss this later.
%nan9: changed according to Ashutosh's suggestion
%For technical reasons, we assume for the next theorem that $|S| = 2$.
%We later discuss how the result must be modified if $|S| = 1$ (as is the
%case in Example~\ref{case:0}).  Intuitively, the theorem says that if
%joe8: 
%For technical reasons, we would consider two seperate cases, for $|S|
%\ge 2$ and $|S| = 1$ (as is the case in Example~\ref{case:0}).  
%joe14: moved this below, and made it a little more palatable
%For technical reasons, we would consider two cases: $|S| \ge 2$
%and $|S| = 1$. 
Intuitively, the theorem says that if
the DM is unaware of some acts that can improve her expected reward, and
the discovery probability is sufficiently low, where ``sufficiently
low'' means $D(1,t) < 1$ for all $t$ and $\sum_{t=1}^\infty D(1,t) <
\infty$, then the DM cannot learn to play near-optimally.  
%Given an MDPU $M = (S, A, S_0, g, g_0,  \A_0, P, R, R^+, R^-, D)$, let
%$M^u_0$ denote the MDP $(S_0,A_0,g_0,(P)_0, R_0)$, where $A_0 = \union_{s
%\in S_0} A_0(s)$, and $(P)_0$ and $R_0$ are the restrictions of $P$
%and $R$ to the actions in $A_0$ and states in $S_0$.  Intuitively,
%$M^u_0$ represents the part of the underlying MDP $M^u$ that the DM is
%aware of.
%joe14: ran on paragraph
%
%nan20: added definition for 'quite knowledgeable', and changed the body
%of t0 and t2 accordingly 
%joe14;  This is good, but we need to give intuition!  Added next sentence
To make the theorem as strong as possible, we show that the lower bound holds
even if the DM has quite a bit of extra information, as characterized
in the following definition.
\begin{definition}\label{d0}
Define a DM to be \textit{quite knowledgeable} if (in addition to $S_0$,
$g_0$, $D$, $R^+$, and $R^-$) she knows $S=S_0$, $|A|$, the transition
function $P_0$, the reward function $R_0$ for states in $S_0$ and
actions in $A_0$, and $R_{\max}$. 
\end{definition}
%joe14
We can now state our theorem.  It turns out that there are slightly
different conditions on the lower bound depending on whether $|S_0| \ge
2$ or $|S_0| = 1$.

%nan7*: body and proof of the theorem rewrote 
\thm \label{t0}
%joe6
%Let $M = (S, A, S_0,  \A_0, P, R, D)$ be an MDPU where $\A_0\subset A$,
%joe7
%Let $M = (S, A, S_0,  \A_0, P, R, R^+, R^-, D)$ be an MDPU 
%where
%nan7: use $D(1,t)$ instead of $f(t)$, and changes made according to comments
%\footnote{$\subset$ here denotes strict subset.}
%joe7: unnecessarily complicated
%there exists a constant $c_1$ such that  $D(1,t)\leq c_1<1$ for all $t$, 
If $D(1,t) <1$ for all $t$ and $\sum_{t=1}^\infty D(1,t)<\infty$, 
%$\A_0\subset A$, 
%%changed $D$ to a more general form
%%joe4*: simplified assumptions on $M$ and weakened assumtpions on f;
%%note that f must be non-negative if D(j,t)
%%\le f, so you don't need to say this.  I don't think we need it to be
%%decreasing.  
%%$D(j,t)\leq f(t)$ for some non-negative decreasing function $f(t)$ where
%%$\sum_{t=1}^{\infty}f(t)<\infty$ and $0<f(1)<1$.  
%there is a function $f$ such that 
%%nan5: i guess we are missing $f(1)<1$ here so as to make $d>0$
%$D(j,t)\leq f(t) \le f(1)<1$ for all $t\ge 1$ and 
%%$D(j,t)\leq f(t) \le f(1)$ for all $t\ge 1$ and 
%$\sum_{t=1}^{\infty}f(t)<\infty$, and
%joe7: moved this up
%MDP $M' = (S,A,P,R)$ is strictly greater than the expected reward of the
%optimal policy in the MDP $(S,A_0,P|_{A_0},R|_{A_0})$ (where
%$P|_{A_0}$ and $R|_{A_0}$ are the restrictions of $P$ and $R$ to
%actions in $A_0$, respectively),
%joe4: much too complicated for a theorem statement
%Let $M'= (S,A,P, R)$ be the underlying MDP; denote the optimal
%expected return for $M'$ to be $r'$. Let $M''= (S, \A_0, P, R)$ be the
%MDP with only initially known actions; denote the optimal expected
%return for $M''$ to be $r''$. Suppose $r'-r''=\alpha>0$.  
%%nan4: inputs added
%The following inputs are given to the DM: $S_0$, $\A_0$, $D$ and $|A|$.
%No algorithm can attain an expected return  $\geq r'-\alpha/2$ on $M$
%with probability greater than $1-c$ for some constant $c\in(0,1)$ in
%finite time.  
%immediate reward for the algorithm is 0
%nan7: changed according to the comments
%joe7
%There exists a constant $c$ such that no algorithm can obtain within
then there exists a constant $c$ such that no algorithm can obtain within
$c$ of the optimal reward for all MDPs that are compatible with what 
%nan20: changed the description of the DM's knowledge to 'quite
%knowledgeable' 
the DM knows, even if the DM is quite knowledgeable,
%the DM knows, even if (in addition to $S_0$, $g_0$, $D$, $R^+$, and $R^-$)
%%joe7
%%the DM knows $S=S_0$, $|A|$ and the maximum possible reward $R_{\max}$.
%the DM knows that $S=S_0$, $|A|$, the transition function $P_0$ and
%reward function $R_0$ for states in $S_0$ and actions in $A_0$, and
%$R_{\max}$, 
%%end nan20
provided that $|S_0| \ge 2$,
$|A| > |A_0|$, 
and $R_{\max}$ is greater than the reward of the optimal policy in the
MDP $(S_0,A_0,P_0,R_0)$.
%joe7*
If $|S_0| = 1$, the same result holds if $\sum_{t=1}^\infty
D(j,t)<\infty$, where $j = |A| - |A_0|$.
\ethm
%joe14
%nan22: we probably need the following paragraph since $M''$ is referred
%later in the proof 
%\fullv{
\prf
%joe4: ``more than finite time'' is not what you want to say.
%We first prove that no matter what algorithm is used, it requires more
%than finite time to discover any new actions with probability greater
%than $c$ for some constant $c\in(0,1)$. 
%nan7: added
%joe7*: largely rewritten.  As stated, this is not OK, because it may be
%that $\A_0(s_1) has more actions than just a_1, so a_1 can't be the
%only action that the DM is unaware of at state s_1.
%We construct an MDP $M''$ that is compatible with what the DM knows,
%such that for $M''$ no algorithm can obtain within $c$ of the optimal reward.
%
%nan7: description of $M''$
%Let $M''=(S, A'', P'', R'')$, let $s_1, s_2\in S$ (this is possible
%since $|S|\ge 2$). Let $a_1$ be the only action that the DM is unaware
%of at state $s_1$; and let $A_2$ be the set of actions that the DM is 
%unaware of at state $s_2$, $|A_2|=|A|-|A_0|-1$; for all other states,
%there are no unknown actions. 
We construct an MDP $M'' = (S,A'',g'',P'',R'')$ that is compatible with what
the DM knows, such that no algorithm can obtain within a constant $c$ of the
optimal reward in $M''$.  The construction is similar in spirit to that
of Example~\ref{case:0}.
%\eprf
%}
%\shortv{We leave details to the full paper. \eprf}
%joe16: removed paragraph break
%
%nan21: recovered the full proof for UAI
%joe14: \end{fullv}
%\fullv{
%nan22: proof now starts from above
%\prf
Since $|S| \ge 2$, let $s_1$ be a 
state in $S$.  Let $j = |A| - |A_0|$, let $A'' = A_0 \union \{a_1,
%nan10: 
%\ldots, a_j\}$, where $a_1, \ldots, a_j$ are fresh actions not in $A$,  
\ldots, a_j\}$, where $a_1, \ldots, a_j$ are fresh actions not in $A_0$,  
let $g''$ be such that  $g''(s_1) = g_0(s_1) \union \{a_1\}$, $g''(s) =
A''$, for $s \ne s_1$.  That is, there
is only one action that the DM is not aware of in state $s_1$, while in
all other states, she is unaware of all actions in $A-A_0$.  
%joe8
%Let $P''(s_1,s_1,a_1)=1$ and $P''(s,s_1,a) = 1$ for all $a \in A''
Let $P''(s_1,s_1,a_1) = P''(s,s_1,a) = 1$ for all $a \in A''
- A_0$ and $s \in S$
(note that $P''$ is determined by $P_0$ in all other cases).
It is easy to check that $M''$ is compatible with what the DM knows,
even if the DM knows that $S=S_0$, knows $|A|$, and knows $R_{\max}$.
%nan9: removed
%Since $R_{\max}$ is greater than the reward of the optimal policy in the
%MDP $(S_0,A_0,P_0,R_0)$, it is clear that the optimal policy in 
Let $R''(s_1,s_1,a_1)= R''(s,s_1,a) = R_{\max}$ for all $s \ne s_1$ and
$a\in A - A_0$  ($R''$ is determined by $R_0$ in all other cases). 
By assumption, the reward of the optimal policy in
%joe9
%$(S_0,A_0,P_0,R_0)$ is less than $R_{\max}$, so the optimal policy is
$(S_0,A_0,g_0,P_0,R_0)$ is less than $R_{\max}$, so the optimal policy is
clearly to get to state $s_1$ and then to play $a_1$ (giving an average
reward of $R_{\max}$ per time unit).  Of course, doing this requires
learning $a_1$. 

As in Example~\ref{case:0}, we first prove that for $M''$ there exists a
constant 
$d > 0$ such that, with probability $d$, no algorithm will discover action
$a_1$ in state $s_1$.
%joe15
%nan22: put this into shortv, so that the full proof in fullv has a
%single exit 
\shortv{
The result then follows as in Example~\ref{case:0}.  We leave details to
the full paper.
\eprf
}

%joe15: cut rest of proof
\fullv{
%As in Example~\ref{case:0}, we first prove that there exists a constant
%$d > 0$ such that, with probability $d$, no algorithm will discover any
%action in $\A - \A_0$.
%joe3*: there's redundancy here; this is the third time you've defined 
%E_t. More importantly, you need to explain the input here.  Does the DM
%know how many actions there are?  If not, this is a weaker result than
%the example you gave before.  Finally, this argument needs to be
%rewritten along the lines of the first example.
%The following steps in (\ref{eq:t2_1}) are the same as what we show in
%(\ref{eq:t2_1}) in Theorem \ref{t2}. 

%nan6: rewrote for notation change
%Thus, we assume without loss of generality
%that $f(1) > 0$. 
%joe6: improved English
%For all state $s$ which has at least one new action to discover (there
%nan7: shorten 
%joe7: shortened further
%Consider $Pr(E_{t,s_1})$. The initial step is exactly the same as in
%Example~\ref{case:0}: 
Again, we have
%For all states $s$ where the DM is unaware of at least one new action
%(there must exist 
%such states since $S_0\subset S$), consider $Pr(E_{t,s})$.
%joe6*: this argument needs to be redone somewhat, to take into account
%more carefully the DM's information, and to get rid of P_{j,s}.  
%The initial steps are exactly the same as in Example~\ref{case:0}:
%\begin{eqnarray}\label{eq:t0_1}	
%nan7: changed to get rid of P_{j,s}
$$\begin{array}{llll}
%joe7
%Pr(E_{t,s})	&=& 	\prod_{t'=1}^t(1-D(1,t')) \\
Pr(E_{t,s_1})	&=& 	\prod_{t'=1}^t(1-D(1,t')).
\end{array}$$
%
%nan7: changed according to the new theorem body
%joe7: proving c_1 exists
Since $\sum_{t=1}^\infty D(1,t) < \infty$, we must have that $\lim_{t
\rightarrow \infty} D(1,t) = 0$.  Since $D(1,t) < 1$ for all $t$, there
must exist a constant $c_1 < 1$ such that $D(1,t) < c_1$ for all $t$.
If $c_1 = 0$, then $D(1,t) = 0$ for all $t \ge 1$
(since $D(1,t) \le c_1$ by assumption, and $D(1,t) \ge 0$,
since it is a probability), so $\Pr(E_{t,s_1}) = 1$, and we can take $d=1$.
%joe7
If $c_1 > 0$, then we show below that $1 - D(1,t') \ge 
(1- c_1)^{D(1,t')/c_1}$.  Thus, we get that  
$$\begin{array}{lll}
\Pr(E_{t,s}) &\ge & \prod_{t'=1}^t (1-c_1)^{D(1,t')/c_1}\\
&\ge &(1-c_1)^{\sum_{t'=1}^t D(1,t')/c_1}.
%\end{eqnarray}			
\end{array}$$
Since, by assumption, $\sum_{t'=1}^\infty D(1,t') < \infty$, we can take 
$d = (1-c_1)^{\sum_{t'=1}^\infty D(1,t')/c_1}$.
%%joe4*: we need this step
%If $f(1) = 0$, then $f(t) = 0$ for all $t \ge 1$
%(since $D(j,t) \le f(t) \le f(1)$ by assumption, and $D(j,t) \ge 0$,
%since it is a probability), so $\Pr(E_{t,s}) = 1$, and we can take $d=1$.
%If $f(1) > 0$, then we show below that $1 - f(t') \ge 
%(1- f(1))^{f(t')/f(1)}$.  Thus, we get that  
%$$\begin{array}{lll}
%\Pr(E_{t,s}) &\ge & \prod_{t'=1}^t (1-f(1))^{f(t')/f(1)}\\
%&\ge &(1-f(1))^{\sum_{t'=1}^t f(t')/f(1)}\\
%&\ge& 	(1-f(1))^{\sum_{t'=1}^\infty f(t')/f(1)}.
%%\end{eqnarray}			
%\end{array}$$
%Since, by assumption, $\sum_{t'=1}^\infty f(t') < \infty$, we can take 
%$d = (1-f(1))^{\sum_{t'=1}^\infty f(t')/f(1)}$.

%joe3* BELOW; YOU NEED TO INCLUDE IT IN THE TEXT.  THIS IS NOT OBVIOUS!]]
%joe4*: rewrote earlier argument, which left out some key steps
%nan7: changed f(t) to D(1,t)
It remains to show that $1-D(1,t') \ge (1-c_1)^{D(1,t')/c_1}$.  
Since $0 \le D(1,t') \le c_1 < 1$, it suffices to show that 
$1-x \ge (1-b)^{x/b} = e^{(x/b)\ln (1-b)}$ for $0 \le x \le b < 1$.
%It remains to show that $1-f(t') \ge (1-f(1))^{f(t')/f(1)}$.  
%Since $0 \le f(t') \le f(1) < 1$, it suffices to show that 
%$1-x \ge (1-c)^{x/c} = e^{(x/c)\ln (1-c)}$ for $0 \le x \le c < 1$.
Let $g(x) = 1-x - e^{(x/b)\ln (1-b)}$.  
We want to show that $g(x) \ge 0$ for $0 \le x \le b < 1$.
An easy substitution shows that 
$g(0) = g(b) = 0$.  Differentiating $g$, we get that 
%nan5: did we miss a 1/c in the second term? please let me know if i'm wrong
%joe5: yes, you're right
%$g'(x) = -1 -\ln(1-c) e^{(x/c)\ln (1-c)}$, and 
$g'(x) = -1 -\frac{\ln(1-b)}{b}~ e^{(x/b)\ln (1-b)}$, and 
%nan5: we probably miss a (1/c)^2 here, please let me know if i'm wrong
%joe5: right again (although it shouldn't affect anything)
%$g''(x) = -e^{(x/c)\ln(1-c)}\ln(1-c)^2 < 0$.  
$g''(x) = -\frac{\ln(1-b)^2}{b^2} ~e^{(x/b)\ln(1-b)} < 0$.  
Since $g(0) = g(b) = 0$ and $g$ is concave, 
we must have  $g(x) = g((1-x/b)0 + (x/b)b) \ge (1-x/b)g(0) + (x/b)g(b) =
0$ for $x \in [0,b]$, as desired.

%nan7: rewrote
%joe9*: Nan, I think r_1 and r_2 are bckwards here
%Let $r_1$ be the expected reward of the optimal policy in 
%$M''$, and let $r_2$ be the expected reward of the optimal
%nan11: you are right
Let $r_2$ be the expected reward of the optimal policy in 
$M''$, and let $r_1$ be the expected reward of the optimal
%joe7
%nan10*: we should use the model with |_{A-\{a_1\}, since we probably
%can easily  get a reward higher than the maximum reward in MDP
%$(S_0,A_0,P_0,R_0)$ (call it r_3) in some cases. 
% E.g. Suppose P(s_1,s_2,a')=1 for some $a'\in A_0$, $s_2\in S_0$ and
% $R(s_1,s_2,a')=r_3$. 
% Thus, once some new action $a''$ is discovered at $s_2$, the agent can play
% $a''$ at $s_2$ to get to $s_1$ and obtain an reward of $R_{\max}$,
% then go back to  
% $s_2$ by playing $a'$ and obtain an reward of $r_3$, and repeat this
% forever. Thus, 
% she gets an expected reward at $(R_{\max}+r_3)/2> r_3$. Moreover, this
% is done without  
% discovering $a_1$ at $s_1$.
policy in the MDP $(S,A''-\{a_1\},P''|_{A-\{a_1\}},R''|_{A-\{a_1\}})$. 
%policy in the MDP $(S_0,A_0,P_0,R_0)$.
%Since $a_1$ is the unique action in $A''$ that achieves the maximum 
%possible reward, therefore $r_1>r_2$.
As we have observed, $r_2 > r_1$.
With probability at least $d$, no algorithm will discover $a_1$, 
%joe7
%so the DM will know only the actions in $A''-\{a_1\}$, and cannot get a
so the DM will know at most the actions in $A''-\{a_1\}$, and cannot get a
reward higher than $r_2$.  Thus, no algorithm can give the DM an
expected reward higher than $(1-d)r_1 + d r_2$.  Thus, we can take 
$c = d(r_1 - r_2)$.  
%%joe4
%%This shows that no algorithm can discover any new action with
%%probability greater than $(1-c)$ no matter how many times $a_0$ is
%%played. 
%Let $r_1$ be the expected reward of the optimal policy in the
%underlying MDP $M'$, and let $r_2$ be the expected reward of the optimal
%policy in the MDP $(S,A_0,P|_{A_0},R|_{A_0})$.  By assumption, $r_1 >
%r_2$.  With probability at least $d$, no algorithm will discover a new
%action, so the DM will know only the actions in $A_0$, and cannot get a
%reward higher than $r_2$.  Thus, no algorithm can give the DM an
%expected reward than $(1-d)r_1 + d r_2$.  Thus, we can take 
%%nan5: a typo here, i think. Let me know if I'm wrong.
%%$c = d(r_1 + r_2)$.  
%$c = d(r_1 - r_2)$.  

%joe7*:
If $|S_0| = 1$, essentially the same argument holds.  We again construct
%joe8
%an MDP $M'' = (S_0,A'',g'',P'',R'')$.  Since $|S_0| = 1$ are
an MDP $M'' = (S_0,A'',g'',P'',R'')$.  Since $|S_0| = 1$, all components
of $M''$ are
determined except for $R''$.  We take $R''(s_1,s_1,a_1) = R_{\max}$, and 
$R''(s_1,s_1,a) = R_{\max} - 1$, for $a \in A'' - (A_0  \union
\{a_1\})$.  Again, the unique optimal policy is to play $a_1$ at all
times, so the problem reduces to learning $a_1$.  Without further
assumptions, all we can say is that this probability of learning $a_1$
after $t$ steps of exploration is at most $D(j,t)$, so we must replace
$D(1,t)$ by $D(j,t)$ in the argument above.% 
%nan9: please check whether the changes I made here are correct 
%joe10: cut for full paper
\fullv{
\footnote{We remark that we can still use $D(1,t)$ if the DM does 
know that $S=S_0$, but does not know $|S|$, and $|S| \ge 2$.  We can also use
%\footnote{We remark that we can still use $D(1,t)$ if the DM does not
%know that $S=S_0$, but does not $|S|$, and $|S| \ge 2$.  We can also use
%joe8: shortened for AAMAS
%$D(1,t)$ if we assume that the probability of learning the specific
%action $a_1$ after $t$ steps of exploration is given by $D(1,t)$.
$D(1,t)$ if the probability of learning the specific
action $a_1$ after $t$ steps of exploration is $D(1,t)$.}}
\eprf
}
%joe11: \end{commentout}
%nan21: end recovered proof

%joe4*: added
%joe7: this wasn't true before with |S| \ge 2, but is now OK
Note that Example~\ref{case:0} is a special case of Theorem~\ref{t0},
since $\sum_{t=1}^\infty \frac{1}{(t+1)^2} < \int_{t=1}^\infty
\frac{1}{t^2} dt = 1$.  
\commentout{
We now consider another instance of
%nan5: I thought of a chess example, let's see if this works
Theorem~\ref{t0}, this one motivated by chess.
In chess, a strategy maps each possible future board position to a legal 
move. Define a winning strategy to be one such that by following 
which the player wins the game no matter what strategy the other 
%joe6*: Nan, I find this a very unreasonable space of actions.  I don't
%think it's reasonable to think of players as trying to find (complete)
%winning strategies (except perhaps late in the game).  In the middle of
%the game, the description of a strategy is huge (since there are a
%large number of board positions).  A player will never be able to
%describe a winning strategy, let alone tell whether a particular
%strategy is a winning strategy.  This means that exploration will be
%pointless.  You could perhaps call an action a partial strategy (one
%that's defined for only a few board positions) that leads to a good
%position according to some evaluation function.  But things get worse
%here.  I think A' should be a fixed set, it certainly doesn't incrase
%exponentially (just as A is a fixed set).  Now, if you assume that the
%number of ``good'' actions grows linearly (or according to some
%low-order polynomial) with the depth, then it follows that the
%probability of finding a good action decreases exponentially.  But this
%is a big assumption.  It seems plausible, but I have no idea if it's
%actually true.  If you're going to use it, it would be nice to have
%some support for it.  (Bart Selman might know about these things.)  If
%you can't find any support, you can write the example under the
%assumption that it's true, but you should make it clear that it's a
%nontrivial assumption.  In %any case, this needs to be rewrittend much
%more carefully, or you need to choose a different example.
player uses. Consider each winning strategy as an action, thus,
$A$ is the set of winning strategies, and $A_0=\emptyset$ originally. 
Define the depth of a 
winning strategy to be the maximum number of moves required for the 
player to win. It is natural to assume that the player evaluates potential 
winning strategies $\A'$ (which is in fact all possible strategies, but
the player may consider it differently) in increasing depth $l$ (here $\A'$ 
includes all potential strategies of depth $\le l$). As $l$ increases, 
$|\A'|$ increases exponentially. Suppose there are only finite number of 
winning strategies, and suppose the player evaluates at most $c$ strategies
at each depth (which is a reasonable assumption for human chess players), 
thus the probability of discovering a winning strategy decreases exponentially 
as time elapses. With these assumptions, the following analysis shows why 
no algorithm finds a winning strategy for chess in finite time.

%In chess, a player has to discover a winning strategy which consists a
%sequence of moves. The player usually considers the potential strategies
%in order of increasing depth. Suppose the player only examines the most
%promising strategies from each depth (which may lead to a winning
%strategy with probability $\rho$ given that the moves from the previous
%depth were chosen correctly). Thus, as depth increases, the probability
%for the player to discover a winning strategy drops exponentially. 
\xam  \label{case:4} %e
%joe4: shortened
%The MDPU in this example is similar to the one in example \ref{case:0},
%but with a different $D$ function.  
%Let $M = (S, A, S_0,  \A_0, P, R, D)$ be an MDPU with 
%$S=S_0=\left\{s_1\right\}$, $\A=\left\{a_1, a_2\right\}$,
%$\A_0=\left\{a_1\right\}$, $P(s_1,s_1,a)=1$ for all action $a$,
%$R(s_1,s_1,a_1)=r_1$ and $R(s_1,s_1,a_2)=r_2$ where $r_2-r_1=\alpha>0$,  
%nan: R(s,s^+,a_0) and R(s,s^-,a_0) added
%$R(s,s^+,a_0)=R(s,s^-,a_0)=0$ 
%and $D(j,t)=\rho^t$
%(The MDPU in this example is the same as in example \ref{case:5}, but
%with a different $D$ function).  
%%inputs:
%The following inputs are given to the DM: $S_0$, $\A_0$, $D$ and
%$|\A|$.  
Consider an MDPU identical to that in Example~\ref{case:0}, except that
$D(t,j) = \rho^t$, where $\rho < 1$.  
%nan7: changed from D(j,t) to D(1,t)
Since $\sum_{t=1}^{\infty}D(1,t)= \sum_{t=1}^{\infty}\rho^t
%Since $\sum_{t=1}^{\infty}D(j,t)= \sum_{t=1}^{\infty}\rho^t
=\frac{\rho}{1-\rho}<\infty$, it follows from Theorem~\ref{t0} that
there is no algorithm that learns to play near-optimally.
\exam
}
In the next section, we show that if $\sum_{t=1}^\infty D(1,t) =
\infty$, then there is an algorithm that learns near-optimal play
(although the algorithm may not be efficient).  
%Theorem~\ref{t0} essentially shows that 
Thus, 
$\sum_{t=1}^\infty D(1,t)$ determines whether or not there
is an algorithm that learns near-optimal play.  
%Essentially, if $\sum_{t=1}^\infty D(1,t) < \infty$, then there is no
%algorithm that will converge to near-optimal play. In the next section
%we will show that if $\sum_{t=1}^\infty D(1,t) = \infty$, then there is
%such an algorithm. 
We can say even more. 
If $\sum_{t=1}^\infty D(1,t) = \infty$, then the efficiency of the best
algorithm for determining near-optimal play depends on how 
quickly $\sum_{t=1}^\infty D(1,t)$ diverges.   Specifically, the
following theorem shows 
that if  
$\sum_{t=1}^T D(1,t) \le  f(T)$, where $f : [1,\infty]
\rightarrow \IR$ is an increasing function whose co-domain includes
$(0,\infty]$ 
(so that $f^{-1}(t)$ is well defined for $t \in (0,\infty]$) and $D(1,t)
\le c < 1$ for all $t$, then the DM cannot learn to play near-optimally with
probability $\ge 1-\delta$ in time less than
%joe9: as you pointed out
%$f^{-1}(\ln(\delta)/\l(1-c))$.  It 
$f^{-1}(c\ln(\delta)/\ln(1-c))$.  It 
follows, for example, that if $f(T) = m_1\log(T) + m_2$, then it requires
time polynomial in $1/\delta$ to learn to play near-optimally with
probability greater than $1-\delta$.  For if $f(T) = m_1\log(T) + m_2$,
then 
%joe9
%$f^{-1}(t) = e^{(t-m_2)/m_1}$, so $f^{-1}(\ln(\delta)/\ln(1-c))
$f^{-1}(t) = e^{(t-m_2)/m_1}$, so $f^{-1}(c \ln(\delta)/\ln(1-c))
= f^{-1}(c\ln(1/\delta)/\ln(1/(1-c)))$ has the form 
$a(1/\delta)^b$ for constants $a, b > 0$.  
%The same lower bound is
%easily seen to hold if $f$ is an increasing function such that $f(t) \le
%m_1\log(t) + m_2$ for all $t$.%
%\footnote{Proof: This follows from the more general claim that if 
%$g$ is an incresing and invertible functions such that $f(x) \le
%g(x)$ for all $x$, then $g^{-1}(x) \le f^{-1}(x)$.  To see this, note
%that $g(g^{-1}(x)) = x = f(f^{-1}(x) \le g(f^{-1}(x))$.  Since $g$ is
%increasing, we must have $g^{-1}(x) \le f^{-1}(x)$.}
%nan10: changed \le to =
A similar argument shows that if $f(T) = m_1 \ln (\ln(T) + 1) +
%A similar argument shows that if $f(T) \le m_1 \ln (\ln(T) + 1) +
%joe9
%m_2$, then $f^{-1}(\ln(1/\delta)/\ln(1/(1-c)))$ has the form 
m_2$, then $f^{-1}(c\ln(1/\delta)/\ln(1/(1-c)))$ has the form 
$ae^{(1/\delta)^b}$ for constants $a, b > 0$; that is, the running time
is exponential in $1/\delta$.  
%joe14
\fullv{
We remark that the assumption that 
$D(1,t) \le c < 1$ for all $t$ is not needed in Theorem~\ref{t0}, since
it already follows from the assumptions that $D(1,t) < 1$ and 
$\sum_{t=1}^\infty D(1,t)<\infty$, since the latter assumption implies
that $\lim_{t \rightarrow \infty} D(1,t) = 0$.}
\thm \label{t2}
%joe7: reworded like previous theorem
%Let $M = (S, A, S_0,  \A_0, P, R, R^+, R^-, D)$ be an MDPU where
%$|S|\ge 2$, $\A_0\subset A$, there exists a constant $c_1$ such that 
%$D(1,t)\leq c_1<1$ for all $t$, $\sum_{t=1}^\infty D(1,t)=\infty$, 
%there exists a constant $m > 0$ such that, for all $T > 0$,
%$\sum_{t=1}^{T}D(1,t)\leq m\ln(\ln T)$, and 
%the expected reward of the optimal policy in the underlying
%MDP $M' = (S,A,P,R)$ is strictly greater than the expected reward of the
%optimal policy in the MDP $(S,A_0,P|_{A_0},R|_{A_0})$.
%For all $\delta>0$, there exists a constant $c$ such that   
%no algorithm can obtain within $c$ of the optimal reward for all
%MDPs that are compatible with what the DM knows in less than 
%exponential time in $\frac{1}{\delta}$, even if (in addition to 
%$S_0$, $g_0$, $D$, $R^+$ and $R^-$) the DM knows $S=S_0$, $|A|$, and  knows
%the maximum possible reward $R_{\max}$.
%joe8*: strengthened assumption; this is necessary
%If $D(1,t) <1$ for all $t$, $\sum_{t=1}^\infty D(1,t) = \infty$, and
%nan20: added 'and' 
%If $\sum_{t=1}^\infty D(1,t) = \infty$, there exists a constant $c < 1$
If 
%joe14: moved up from below
$|S_0| \ge 2$, $|A| > |A_0|$, $R_{\max}$ is greater than the reward of
the optimal policy in the MDP $(S_0,A_0,P_0,R_0)$, 
$\sum_{t=1}^\infty D(1,t) = \infty$, and there exists a constant $c < 1$
%nan20: changed '<' to '\le', as we described in the paragraph before the theorem
%such that $D(1,t) < c$ for all $t$, and
such that $D(1,t) \le c$ for all $t$, and
%nan9: we need m_2 here, since \ln(\ln 1) and \ln(\ln 2) are invalid or
%negative values, 
% actually the most general form should be O(\ln(\ln T)).
%joe8*: gave a more abstract statement, which allowed me to simplify
%things (getting rid of a few constants)
%> 1; as you point out, \ln\ln(1) and \ln\ln(2) are undefined
%there exists constants $m > 0$ and $m_2$ such that, for all $T > 0$,
%$\sum_{t=1}^{T}D(1,t)\leq m\ln(\ln T)+m_2$,
an increasing function $f: [1,\infty] \rightarrow \IR$ such that the
co-domain of $f$ includes $(0,\infty]$ and 
%there exists a constant $m > 0$ such that, for all $T > 0$,
%$\sum_{t=1}^{T}D(1,t)\leq m\ln(\ln T)$,
$\sum_{t=1}^{T}D(1,t)\leq f(T)$,
%joe8*: need to add \delta here
then for all $\delta$ with $0 < \delta < 1$,
there exists a constant $d > 0$ such that no algorithm that runs in time
%joe8*: more general statement.  
%time less than exponential in $\frac{1}{\delta}$
%joe9
%less than $f^{-1}(\ln(\delta)/\ln(1-c))$ 
less than $f^{-1}(c\ln(\delta)/\ln(1-c))$ 
can obtain within
$d$ of the optimal reward for all MDPs that are compatible with what 
the DM knows
%joe8*: we need this
with probability $\ge 1-\delta$,
%nan20: changed the description of the DM's knowledge to 'quite knowledgeable' (defined in definition \cite{d0})
even if the DM is quite knowledgeable.
%even if (in addition to $S_0$, $g_0$, $D$, $R^+$, and $R^-$)
%the DM knows that $S=S_0$, $|A|$, the transition function $P_0$ and
%reward function $R_0$ for states in $S_0$ and actions in $A_0$, and
%$R_{\max}$, 
%%end nan20
%joe14: moved this back; I think that makes it easier to parse.
%provided that $|S_0| \ge 2$, $|A| > |A_0|$, 
%and $R_{\max}$ is greater than the reward of the optimal policy in the
%MDP $(S_0,A_0,P_0,R_0)$.
%nan10: fixed typo
If $|S_0| = 1$, the same result holds if $\sum_{t=1}^T
D(j,t)\le f(T)$, where $j = |A| - |A_0|$.
%If $|S_0| = 1$, the same result holds if $\sum_{t=1}^\infty
%D(j,t)<\infty$, where $j = |A| - |A_0|$.
\ethm
%joe11: 
\fullv{
\prf
%joe8
%Construct the same $M''$ as the one we constructed in Theorem
%\ref{t0}. It is easy to check that $M''$ is compatible with
Consider the MDP $M''$ constructed in the proof of Theorem \ref{t0}. 
As we observed, $M''$ is compatible with
%joe8: added comma
%what the DM knows (even if the DM knows $S=S_0$, $|A|$ and the maximum
what the DM knows (even if the DM knows $S=S_0$, $|A|$, and the maximum
possible reward $R_{\max}$). 
%We prove that for $M''$ no algorithm can obtain within $c$ of the
%optimal reward in less than exponential time in $\frac{1}{\delta}$. 
%joe6*: This proof will have to be rewritten slightly to take into
%account the DM's uncertainty, like the earlier impossibility result.
%joe5
%We first prove for any $\delta>0$, no matter what algorithm is used, it
%requires at least time exponential in $\frac{1}{\delta}$ to discover
%any
%joe7*: added
%joe8: another comma
%Note that, for all $\epsilon > 0$ the $\epsilon$-return mixing time of
Note that, for all $\epsilon > 0$, the $\epsilon$-return mixing time of
%nan9: we have an incomplete sentence here
%joe8
%$M''$ is 1..
$M''$ is 1.
%nan10: I see what's going on now. :)
%joe9*: Nan, if it took you a while to figure this out, you may want to
%add a few sentences to make it easier for other readers to understand.
%nan11: we are probably fine. 

We now prove for all $\delta>0$, all algorithms 
%nan7: 
%joe8*
%require at least time exponential in $\frac{1}{\delta}$ to discover $a_1$
require at least time $f^{-1}(\ln(\delta)/\ln(1-c))$ to discover $a_1$
in $M''$ with probability $\geq 1-\delta$.  
%require at least time exponential in $\frac{1}{\delta}$ to discover a
%new action with probability $\geq 1-\delta$.  
%nan7: rewrote with $f(t)$ replaced by $D(1,t)$
%joe7:  
%nan9: we would need the following argument to show that there exists a
%$c_1$ where $D(1,t) < c_1$ for all $t$ 
%joe8*: This argument is false, unfortunately.  For a counterexample,
%suppose that $D(1,e^{e^k}) = 1-1/k$, and $D(1,t) = 0$ if $t$ does not have
%the form $e^{e^k}$.  Then it is easy to see that \sum_{t=1}^{e^{e^k}}
%\le k, so this satisfies the hypotheses of the theorem, although D has
%an increasing subsequence.  I modified the theorem's assumptions to
%take this into account
%Since $\sum_{t=1}^{T}D(1,t)\leq m\ln(\ln T)+m_2$ for all $T$,
%differentiating $m\ln(\ln T)$ gives $\frac{m}{(\ln T)T}$ which is
%strictly decreasing, thus $m\ln(\ln T)$ is concave. This shows that
%$D(1,t)$ cannot contain an infinite series of increasing values.  
%Moreover, since $D(1,t) < 1$ for all $t$, there
%must exist a constant $c_1 < 1$ such that $D(1,t) < c_1$ for all $t$.
By assumption, there exists a constant $c < 1$ such that $D(1,t) <
c$ for all $t$. 
%joe8
%If $c = 0$, then $D(1,t) = 0$ for all $t \ge 1$ (since $D(1,t) \le
%c$ by assumption, and $D(1,t) \ge 0$, since it is a
%probability), so 
We must have $c > 0$, for otherwise $D(1,t) = 0$ for all $t$ and 
$\sum_{t=1}^{\infty}D(1,t)=0\neq \infty$, a
contradiction. 
%joe8
%Thus $c_1 > 0$. 
The same argument as in Theorem \ref{t0} now shows that 
%joe8: changed c_1 -> c
$\Pr(E_{t,s_1}) \ge (1-c)^{\sum_{t'=1}^t D(1,t')/c}$.  Since 
%nan9: changes made according to m_2
%$\sum_{t'=1}^t D(1,t') \le m\ln (\ln t)$, it follows that 
%$\Pr(E_{t,s_1}) \ge (1-c_1)^{m\ln (\ln t)/c_1}$.
%joe8: here things simplify, because I don't care about the constants
%$\sum_{t'=1}^t D(1,t') \le m\ln (\ln t)+m_2$, it follows that 
%$\Pr(E_{t,s_1}) \ge (1-c_1)^{(m\ln (\ln t)+m_2)/c_1}$.
$\sum_{t'=1}^t D(1,t') \le f(t)$, it follows that 
%nan10*: we probably need a 1/c, it is added here after
$\Pr(E_{t,s_1}) \ge (1-c)^{f(t)/c}$.
%$\Pr(E_{t,s_1}) \ge (1-c_1)^{f(t)}$.
Note that for the probability of discovering $a_1$ to be at least
$1-\delta$ at state $s_1$, we must have $\Pr(E_{t,s_1}) \le \delta$, which
in turn requires that 
%nan9: changes made according to m_2
%$(1-c_1)^{m\ln (\ln t)/c_1}\le\delta$. 
%joe8: 
%$(1-c_1)^{(m\ln (\ln t)+m_2)/c_1}\le\delta$. 
$(1-c)^{f(t)/c}\le\delta$. 
Taking logs of both sides and rearranging terms, 
%joe8
%using the fact that $\ln(1-\delta) < 0$, we must have $\ln (\ln(t)) \ge (c_1
%nan10*: would it be \delta instead of 1-\delta? changes are made here after
we must have $f(t) \ge c\ln(\delta)/\ln(1-c)$, so $t \ge
f^{-1}(c\ln(\delta)/\ln(1-c))$, since $f$ is increasing. (Note that
%we must have $f(t) \ge \ln(1-\delta)/\ln(1-c)$, so $t \ge
%f^{-1}(\ln(1-\delta)/\ln(1-c))$, since $f$ is increasing. (Note that
since $0 < \delta < 1$ 
and $0 < c < 1$, both $\ln(1-c)$ and $\ln(\delta)$ are negative, so 
$\ln(\delta)/\ln(1-c) > 0$, and $f^{-1}(c\ln(\delta)/\ln(1-c))$
is well defined.)
%nan9: changes made according to m_2
%ln(\delta))/(m\ln(1-c_1)) = (c_1 \ln (1/\delta)) (m \ln(1/(1-c_1)))$.
%Thus $t \ge e^{(c'/\delta)}$, where $c' = c_1/(m \ln(1/(1-c_1)))$ is a
%joe8: unnecessary now
%ln(\delta))/(m\ln(1-c_1))-\frac{m_2}{m} = (c_1 \ln (1/\delta))/(m
%\ln(1/(1-c_1)))-\frac{m_2}{m} $. 
%Thus $t \ge e^{c''(1/\delta)^c'}$, where $c' = c_1/(m \ln(1/(1-c_1)))$
%and $c''=e^{-m_2/m}$ are  
%positive constants.
%joe8
%
%Thus, it requires at least time exponential in $\frac{1}{\delta}$ 
%nan10: added c
Thus, it requires at least time $f^{-1}(c\ln(\delta)/\ln(1-c))$
%Thus, it requires at least time $f^{-1}(\ln(\delta)/\ln(1-c))$
to discover $a_1$ with probability $\geq 1-\delta$.

Let $r_1$ be the expected reward of the optimal policy in $M''$, 
and let $r_2$ be the expected reward of the optimal
policy in the MDP $(S,A'',P''|_{A''-\{a_1\}},R''|_{A''-\{a_1\}})$.  By
the construction of $M''$, $r_1 > 
r_2$. If $a_1$ is not discovered, the DM will know only the actions in $A''-\{a_1\}$, and cannot get a
reward higher than $r_2$.  It follows that no algorithm can give the DM an
%joe8
%expected reward greater than $(1-\delta)r_1 + \delta r_2$ in less than time 
expected reward greater than $d = (1-\delta)r_1 + \delta r_2$ in time 
%nan10: added c
less than $f^{-1}(c\ln(\delta)/\ln(1-c))$.
%less than $f^{-1}(\ln(\delta)/\ln(1-c))$.
%%nan6: slightly rewrote
%Let $r_1$ be the expected reward of the optimal policy in the
%underlying MDP $M'$, and let $r_2$ be the expected reward of the optimal
%policy in the MDP $(S,A_0,P|_{A_0},R|_{A_0})$.  By assumption, $r_1 >
%r_2$. If no new action is discovered, the DM will know only the actions
%in $A_0$, and cannot get a 
%reward higher than $r_2$.  It follows that no algorithm can give the DM an
%expected reward greater than $(1-\delta)r_1 + \delta r_2$ in less than time 
%%joe6
%%exponential in $\frac{1}{\delta}$.  Thus, we can take $c = d(r_1 -
%%r_2)$.  
%exponential in $\frac{1}{\delta}$.  Thus, we can take $c = \delta(r_1 -
%r_2)$.   
\eprf
}
%joe11: \end{commentout}
%end nan4

%joe6*: added.  This is an important part of the story.
%joe7
%It is not hard to show that if $\sum_{t=1}^T D(1,t) \ge m \ln (\ln t)$
%joe8: shortened 
%Again, in the next section, we prove that the lower bound of
\begin{sloppypar}
In the next section, we prove that the lower bound of
%joe9
\shortv{Theorem~\ref{t2} is tight.}
%joe14:
\fullv{
Theorem~\ref{t2} is essentially tight:
%joe8
%if $\sum_{t=1}^T D(1,t) \ge m \ln (\ln t)$
if $\sum_{t=1}^T D(1,t) \ge f(T)$, then the DM can learn to play
%nan10*: we don't need c when proving upper bounds, since we direct have
%1-D(1,t)\ge e^{-D(1,t)} 
% I think we would also need 4N here, so I added it. But now the upper bound and lower bound
% formulas don't match as well as before. Or maybe we are ok here without the 4N...?
near-optimally in time polynomial in \mbox{$f^{-1}(\ln(4N/\delta))$}
%near-optimally in time polynomial in \mbox{$f^{-1}(\ln(\delta)/\ln(1-c))$}
and all the other parameters of interest.  In particular, if 
$f(t) \ge m_1 \ln(t) + m_2$ for some constants $m_1$ and $m_2$, then
the DM can learn to play near-optimally in time polynomial in the
relevant parameters.
%for some constant $m > 0$.
%%joe7
%then we can find a near-optimal algorithm in exponential time.
%then the DM can learn to play near-optimally in exponential time.
%joe8
%Of course, the question of most interest is what we need to assume about
%$\sum_{t} D(1,t)$ to get a polynomial time algorithm.  As we shall see
%in the next section, we can get polynomial time if we replace $\ln (\ln
%t)$ by $\ln t$.
}
\end{sloppypar}

%joe7*: this material should move to the upper bound section, and needs
%to be rewritten in line with our current notation.  We should describe
%what the DM knows, and then point out that the family of MDPs (not
%MDPUs) are all compatible with the DM's knowledge.  I cut it for now;
%it's an obvious candidate for cutting from the submission
%We conclude this observation regarding the importance
\fullv{
We conclude this section with an observation regarding the importance
of the DM's information.  
%joe7
%All our impossibility results hold even if the
As we have seen,
Theorems~\ref{t0} and \ref{t2} hold even
DM has a great deal of information; specifically, the DM can know 
that $S=S_0$, $|A|$,  and the maximum possible reward.  On the other hand, our
near-optimal policy construction does not require this information.  It
turns out that the assumption
that $D(j,t) \ge D(1,t)$ plays a crucial role in these results.  While
the assumption seems quite natural (it 
would be strange if the probability of discovering a new action
decreased as the number of undiscovered actions increases), it is worth
noting that without this assumption, there are cases where the DM can
find a near-optimal policy if she knows $|A|$, but does not know it
otherwise.

%[[NAN, IF YOU AGREE WITH MY COMMENTS, I WOULD GO DIRECTLY TO
%EXAMPLE 3.9 HERE (SUITABLY REWRITTEN), AND CUT THEOREMS 3.6--3.8.]]
%nan7: I agree with you

%[[NAN, YOU NEED TO SLOW DOWN AND EXPLAIN WHAT IT MEANS NOT TO KNOW
%$|A|$.  THIS EXPLANATION MAY NEED TO GO EARLIER, IF IT'S RELEVANT TO THE
%THEOREM THAT YOU ARE ADDING.]]
%nan6: added explanation before Theorem \ref{t1}. 
%joe5*: 
%[[HERE: I AM HOPING THAT YOU'll PROVE A MORE GENERAL RESULT ABOUT WHAT
%HAPPENS IF WE CAN'T FIND AN $f(t)$ THAT BOUNDS $D(j,t)$ INDEPENDENT OF $j$]]
%nan6*: new theorems added below :) 

%In all the above examples, not knowing whether you have discovered all
%actions does not affect the lower bound. We now give an example where it
%does make a difference.  

%nan4*: new example added for the case where whether knowing an upper bound of the number of actions makes a big difference
\xam \label{case:6}
A family of MDPUs $M^*=\{M^1, M^2, M^3, \cdots, M^{\infty}\}$ are involved in this example, 
where $M^i=(S, \A^i, S_0, \A_0, P, R^i, R^+, R^-, D)$ such that $S=S_0=\{s_1\}$, $\A^i=\{a_1,a_2,\cdots,a_i\}$,
$\A_0=\emptyset$, $P(s_1,s_1,a)=1$ for all action $a$, $R^i(s_1,s_1,a_j)=j$ where $j\in\{1,2,\cdots,i\}$,
%nan4
$R^+(s,a_0)=R^-(s,a_0)=0$ for all state $s$
and $D(j,t)=\frac{1}{t+j}$. Note that the same $S, S_0,\A_0,P$ and $D$ are shared across the family. 
The DM is in one of the MDPUs among the $M^*$ family.
%nan5: cut
%, and he is given the following inputs: $S_0$, $\A_0$ and $D$. 
The DM knows that she is in a member of the $M^*$ family, but she does not know which. We now show that whether an upper bound of the  number of actions being given to the DM makes a big difference to her performance.

%nan7: changed to the previous version of proof with some revision
%%nan6*: proof rewrote according to the new theorems
%%joe6*: Nan, this is the wrong story.  If the DM konws |A|, then she
%%doesn't consider all the MDPs in M* possible; she knows that M^k for k >
%%|A| is impossible
%First, assume that the DM knows $|A|$. Take $f(t)=\frac{1}{|A|+t}$, and the problem becomes a special case of Theorem \ref{t5}. Thus, there exists an algorithm which learns near-optimal play. (In fact, there are algorithms which learns near-optimal play in polynomial time.)
%
%Now suppose the DM does not know $|A|$. Since $D(j,t)=\frac{1}{t+j}\le 1/2<1$ for all $t$ and $j$, and for any $t>0$, we can choose $j=t$, and thus $\sum_{t'=1}^t D(j,t')= \sum_{t'=1}^t \frac{1}{t'+t}= \frac{1}{1+t}+\frac{1}{2+t}+\cdots+\frac{1}{t+t}\le t\cdot\frac{1}{1+t}\le 1$, therefore, this is a special case of Theorem \ref{t6}, and no algorithm may learn to play near-optimally.
%

%nan6: cut
%nan6: changed according to new notation: E_{t,s}
%nan7: rewrote
For all state $s$ that has at least one new action, assume there are $j_s$ 
actions unaware to the DM (note that the DM does not know the value of $j_s$), we have
%For all state $s$ that has at least one new action, we have
%According to the definition of $E_{t,s}$, we have
%nan5: changed to array
%			\begin{eqnarray}\label{eq:e6_0}	
$$\begin{array}{lll}
%nan7: rewrote to remove $P_{j,s}$
%				Pr(E_{t,s})	&=& \sum_{j>0} P_{j,s}\cdot\prod_{t'=1}^t(1-\frac{1}{j+t})\nonumber \\
%								&=& \sum_{j>0} P_{j,s}\cdot(\frac{j}{j+1}\times\frac{j+1}{j+2}\times\frac{j+2}{j+3}\times\cdots
%										\frac{j+t-1}{j+t})\nonumber \\
%								&=& \sum_{j>0} P_{j,s}\cdot\frac{j}{t+j}.				
				Pr(E_{t,s})	&=& \prod_{t'=1}^t(1-\frac{1}{j_s+t})\nonumber \\
								&=& (\frac{j_s}{j_s+1}\times\frac{j_s+1}{j_s+2}\times\frac{j_s+2}{j_s+3}\times\cdots
										\frac{j_s+t-1}{j_s+t})\nonumber \\
								&=& \frac{j_s}{t+j_s}.				
\end{array}$$
%			\end{eqnarray}			
			
First, assume that the DM knows $k$ -- an upper bound on the actual number of actions in the underlying MDPU. Thus
%nan7:
$k\ge j_s$, and 
%nan5: changed to array
			%\begin{eqnarray}\label{eq:e6_1}	
$$\begin{array}{llll}			
%nan7: changed to remove $P_{j,s}$
%				\Pr(E_{t,s})	&=& 	\sum_{j=1}^k P_{j,s}\cdot\frac{j}{t+j}\nonumber\\
%								&\leq&	\sum_{j=1}^k P_{j,s}\cdot\frac{k}{t+k}\nonumber\\
%								&\leq& \frac{k}{t+k}&\mbox{[since $\sum_{j=1}^k P_{j,s} = 1$]}.
				\Pr(E_{t,s})	&\leq& \frac{k}{t_s+k}.
\end{array}$$
%			\end{eqnarray}			

%nan5: shorten	
%	Set (\ref{eq:e6_1}) to be no greater than $\delta$, We get 
%	\[t\geq \frac{(1-\delta)k}{\delta}.\]
Take $t= \frac{(1-\delta)k}{\delta}$, we get $\Pr(E_{t,s})\leq\delta$.
	Thus, if the DM knows $k$, she is guaranteed to discover all actions with probability $\geq (1-\delta)$ for any $\delta>0$ in time polynomial in $k$ and $1/\delta$ no matter which game she is playing. It follows easily that she is also guaranteed to achieve the optimal reward with probability $\geq (1-\delta)$ for any $\delta>0$ in polynomial time.
	
	Now suppose the DM does not know $k$. We shall prove that the DM cannot obtain an optimal reward no matter how many times $a_0$ is played. Suppose the highest reward she could get from any currently discovered action is $r$, and suppose she has played $a_0$ for $t>0$ times without finding any new actions. Since the underlying MDPU can be any member of the $M^*$ family, let it be $M^i$ where $i=\max(9t,10r)$. Thus, the optimal reward is $i\geq 10r$, which is 10 times of what the DM has currently achieved.

	We now show that this actually happens with constant probability. %By (\ref{eq:e6_0}),
%nan5: changed to array
$$\begin{array}{llll}			
			Pr(E_{t,s}) &=&		\frac{i}{t+i} \nonumber\\
							&\geq&	\frac{9t}{t+9t} \nonumber\\
							&=&			\frac{9}{10}. \nonumber\\							
\end{array}$$
	In this case with probability $\geq \frac{9}{10}$ the DM only attains $\frac{1}{10}$ of the optimal reward.
	In fact, we can set $i=\max(nt,~nr+r)$ with arbitrarily large
$n$, in which case with probability $\geq \frac{n}{n+1}$ the DM
only attains $\frac{1}{n+1}$ of the optimal reward. 
	
	In conclusion, if the DM does not know $k$, she can attain an
arbitrarily low reward compared to the optimal reward no matter how many
times she plays $a_0$. On the other hand, if she knows $k$, she is
guaranteed to achieve a near-optimal reward in polynomial time. 
%nan22: added \eprf
\eprf
\exam
}
%joe7:\end{fullv}

%\section{Learning to play near-optimally}\label{sec:urmax} 
%joe16
%\section{LEARNING TO PLAY NEAR-OPTIMALLY}\label{sec:urmax} 
\section{\mbox{LEARNING TO PLAY NEAR-OPTIMALLY}}\label{sec:urmax} 
In this section, we show that a DM can learn to play near-optimally in
%nan21: changed 'with' to 'where'
%an MDPU with $\sum_{t=1}^\infty D(1,t)=\infty$. Moreover, we show that when 
an MDPU where $\sum_{t=1}^\infty D(1,t)=\infty$. Moreover, we show that when 
$\sum_{t=1}^\infty D(1,t)=\infty$, the speed at which $D(1,t)$ decreases 
determines how quickly the DM can learn to play near-optimally.
%joe8: to save space
\fullv{Specifically, 
if $\sum_{t=0}^T D(1,t)\ge m_1 f (\ln T) + m_2$
for all $T>0$ for constant $m_1>0$ and $m_2$, and an invertible function
$f$, then the DM can learn to play near-optimally in time polynomial in
$f^{-1}(1/\delta)$.  In particular, if $f$ is the identity (so that 
$\sum_{t=0}^T D(1,t)\ge m_1 \ln T + m_2$), then 
the DM can learn in time polynomial in $1/\delta$ (and, as we shall see,
in time polynomial in all other parameters of interest).
}
%joe12
\fullv{
We call the learning algorithm $\URMAX$, since it is an extension of
$\RMAX$ to MDPUs.} 
%joe12*: In response to Ashutosh's comments
While the condition $\sum_{t=1}^\infty D(1,t)=\infty$ may seem rather
special, in fact it arises in many applications of interest.  
For example, when learning to fly a helicopter 
\cite{AN05ICML,SSS09}, the space of potential actions in which the exploration
takes place, while four dimensional (resulting from the six degree of
freedom of the helicopter), can be discretized and taken to be finite.
%joe14*: Nan, this is not obvious at all.  It depends on how exploration
%is done!  This *must* be clarified.
%nan21*: added the method of exploration below
%joe15
%Thus, if our procedure for exploration is by examining the potential 
%actions uniformly at random, then $D(1,t)$ is constant for all t, 
Thus, if we explore by examining the potential 
actions uniformly at random, then $D(1,t)$ is constant for all $t$, 
%Thus, $D(1,t)$ is constant for all $t$, 
and hence $\sum_{t=1}^\infty
D(1,t)=\infty$.  Indeed, in this case $\sum_{t=1}^T
D(1,t)$ is $O(T)$,   so it follows from Corollary~\ref{cor:poly} below
that we can learn to fly the helicopter near-optimally in polynomial
time.  The same is true in any situation where the space of potential
actions in which the exploration takes place is finite and understood.

%joe11
%$\URMAX$ is an extension of 
%(Brafman and Tennenhotlz, 2002)'s $\RMAX$ algorithm as described in
%Brafman and Tennenhotlz's \citeyear{BT02} $\RMAX$ algorithm as described in
%Section \ref{sec:pre}. 

%nan7*:rewrote
%joe7: rewrote and added subsection heading
%joe14: No need to have a single subsection in a section
%\subsection{The $\URMAX$ algorithm }
%joe8:
We assume throughout this section that $\sum_{t=1}^\infty D(1,t) = \infty$.
%R-max does not apply to MDPU problems since the basic assumption that 
%all actions are known is broken.
%joe8: added
We would like to use an $\RMAX$-like algorithm to learn to play
near-optimally in our setting too, but there 
are two major problems in doing so.
%nan9
%joe8: cut this
%We must modify to deal with MDPUs.
%We must modify to deal with MDPUs, in our to take into account the fact that 
%joe7*: I don't think we need this
%nan7*
%The first is that, unlike R-Max,  we do not assume that the DM knows all
%the actions. We deal with this problem by assuming that the DM has an 
%upper bound on the number of actions.  We also assume that the DM has an
%upper bound on the number of states. All our
%The first is that, unlike R-Max,  we do not assume that all
%states and actions are known. We deal with this problem by assuming that
%we have an upper bound on the number of states and actions.  All our
%results are stated in terms of this upper bound; that is, our polynomial
%time results are polynomial in the bound on states and actions, not in
%the actual number of states and actions. 
The first is that we do not want to assume that the DM knows $|S|$,
$|A|$, or $R_{\max}$.  We deal with the fact that $|S|$ and $|A|$ are
unknown by using essentially the same idea as Kearns and Singh use for
dealing with the fact that the true $\epsilon$-mixing time $T$ is
unknown: we start with an estimate of the value of $|S|$ and $|A|$, and
keep increasing the estimate.  Eventually, we get to the right values,
and we can compensate for the fact that the payoff return may have been
too low up to that point by playing the policy sufficiently often.
The idea for dealing with the fact that $R_{\max}$ is not known is
similar.  We start with an estimate of the value of $R_{\max}$, and 
recompute the value of $K_1(T)$ and the approximating MDP every time we
discover a transition with a reward higher than the current estimate.
%joe8
%(We remark that this idea can be applied to $R-max$ as well.)
(We remark that this idea can be applied to $\RMAX$ as well.)
%joe7
%%nan7: added
%We will show later that this assumptions can be removed.
%Since we assume that,
%typically, the DM has a reasonable estimate of the number of states and
%actions, we do not view this as a major issue. The key difficulty
%joe14: ran on paragraph
%
The second problem is more serious: we need to deal with the fact that
not all actions are known, 
%joe7
and that we have a special \emph{explore} action.
%The key difficulty 
%in applying the R-max ideas to our setting is in dealing with the
%special explore action.  
Specifically, we need to come up with an analogue 
%joe7
%of $K$ that takes the \emph{explore} action into account.
%Previously, R-max used $K_1$  
of $K_1(T)$ that describes how many times we should play the explore action
%nan10: typo 
$a_0$ in a state $s$, with a goal of discovering all the actions in $s$.
%$a_0$ in a state $s$, with a goal of discovering all the actions in $a$.
%joe14
\fullv{
Clearly this value will depend on the discovery probability (it turns out
that all we need to know is $D(1,t)$ for all $t$) in addition to all the
parameters that $K_1(T)$ depends on.
}

We now describe the $\URMAX$ algorithm under the assumption that the DM
knows $N$, an upper bound on the state space $S$, $k$, 
an upper bound on the size of the action space $A$, $R_{\max}$,
an upper bound on the true maximum reward, and $T$, an upper bound
on the $\epsilon$-return mixing time.  To emphasize the dependence on
these parameters, we denote the algorithm
$\URMAX(S_0, g_0, D, N, k, R_{\max},T,\epsilon,\delta,s_0)$.  (The DM may
also know $R^+$ and $R^-$, but the algorithm does not need these
inputs.)  We later show how to define $\URMAX(S_0, g_0, D,
%nan10: typo
\epsilon,\delta,s_0)$, dropping the assumption that the DM knows $N$,
$k$, $T$ and $R_{\max}$.
%\epsilon,\delta,s_0)$, dropping the assumption that the DM knows $D$, $N$,
%$k$, and $R_{\max}$.

Define
%new $K_1$
\begin{itemize}
%joe10
\denselist
%joe8
%\item Define $K_1 =
\item $K_1(T) =
%nan10: should be 4N, the same as in $\RMAX$. also, since we have a single
%starting state $s_0$ now, 
% we only need T instead of T+1 (previously we had $T+1$ because of assuming an initial distribution)
\max((\lceil\frac{4NTR_{\max}}{\epsilon}\rceil)^3,$ $\lceil
%\max((\lceil\frac{8N(T+1)R_{\max}}{\epsilon}\rceil)^3,$ $\lceil
%joe9: I tink it's better to have the delta in denominator and get rid
%of the -
%-8\ln^3(\frac{\delta }{8Nk})\rceil)+1$;
8\ln^3(\frac{8Nk}{\delta })\rceil)+1$;
%joe8: moved below
%\footnote{The new $K_1$ value has a coefficient 8 instead of 6 due to
%the new possibility that an MDPU algorithm may fail because not all
%joe8*: I don't think the DM has to know all the states for BT either,
%so this needs to be checked.  BT do need an upper bound on the number
%of states, but we have that too for this part of the algorithm.
%actions are discovered. Since the DM no longer knows all the states,
%$K_1$ also has an extra coefficient $T+1$. Besides that, it has one less
%$k$ (that is $|A|$ in $\RMAX$) in every expression, because unlike SGs,
%MDPs are single player games and the DM does not have to consider the
%component's actions.} %end of footnote 
%define $K_0$
%joe8
%\item Define $K_0$ to be the least $M$ such that $\sum_{t=1}^M D(1,t)\ge
%\ln(4N/\delta)$, and we take a state-action pair $(s,a_0)$ to be known
%if it is played $K_0$ times (rather than $K_1$ times). 
%joe11: there's something wrong here
%\item $K_0 = \mathrm{argmin}_{M} = \sum_{t=1}^M D(1,t)\ge
%\ln(4N/\delta)$.
\item $K_0 = \min_M\{M: \sum_{t=1}^M D(1,t)\ge \ln(4N/\delta)\}$.
(Such a $K_0$ always exists if $\sum_{t=1}^M D(1,t) = \infty$.)
\end{itemize}
Just as with $\RMAX$, $K_1(T)$ is a bound on how long the DM needs to get a
good estimate of the transition probabilities at each state $s$.  Our
definition of $K_1(T)$ differs slightly from that of Brafman and Tennenholtz
%nan10: removed 'extra term $T+1$', since we are using T now just as in
%R-max 
(we have a coefficient 8 rather than 6; the
%(we have a coefficient 8 rather than 6, and an extra term $T+1$; these
difference turn out to be needed to allow for the fact that we do not
know all the actions).  As we show below (Lemma~\ref{l0}), $K_0$ is a good
estimate on how often the \emph{explore} action needs to be played in
order to ensure that, with high probability (greater than
%joe11*: this is not what the lemma proves, as an AAMAS reviewer
%correctly pointed out
%$1-\delta/4N$), all actions are discovered at a state.
$1-\delta/4N$), at least one new action is discovered at a state, if
there is a new action to be discovered.
Just as with $\RMAX$, we take a pair $(s,a)$ for $a \ne a_0$ to be
\emph{known} if it is played $K_1$ times; we take a pair $(s,a_0)$ to be
\emph{known} if it is played $K_0$ times.

%joe8
%And the $\URMAX$ proceeds just like $\RMAX$ except that: 
$\URMAX(S_0, g_0, D, N, k, R_{\max},T,\epsilon,\delta,s_0)$ proceeds
just like $\RMAX(N, k, R_{\max},T,\epsilon,\delta,s_0)$, except for the
following modifications:
\begin{itemize}
%joe10
\denselist
%quit and restart $\URMAX$ with the new values of $R_{\max}$, $k$ or $N$
%joe8
%\item we will quit if we discover a reward $> R_{\max}$,  more than $k$
\item The algorithm terminates if it discovers a reward greater than
$R_{\max}$,  more than $k$ actions, or more than $N$ states 
($N$, $k$, and $R_{\max}$ can be viewed as the current guesses for these
values; if the guess is discovered to be incorrect, the algorithm is
restarted with better guesses.)
%nan9*: we can probably use $R^-(s)$ instead of $-\infty$,
%because if all other actions give reward lower than $R^-(s)$,
%then the DM should keep playing $a_0$ at state $s$ although she
%does not discover any new action at $s$ any more. 
\item if $(s,a_0)$ has just become known, then we set the reward for
%joe8*: we can use -\infty, since this means we don't have to know R^-(s)
%playing $a_0$ in state $s$ to be $R^-(s)$ - the true reward for playing
playing $a_0$ in state $s$ to be $-\infty$ (this ensures that $a_0$ is
not played any more in state $s$).
\end{itemize}
%joe8: added
%joe14: 
%We say that \emph{an inconsistency is discovered} if the algorithm
For future reference, we say that \emph{an inconsistency is discovered}
if the algorithm terminates because it discovers a reward greater than
$R_{\max}$,  more than $k$ actions, or more than $N$ states. 

%joe8
%We will prove results showing that if the estimates on the input
%parameters are correct, then $\URMAX$ has good performance. Before that,
%we first explain our choice of $K_0$. This $K_0$ guarantees that if
%there is a new action to be discovered, then with probability $\ge
%1-\delta/4N$ it will be discovered. 
%joe14
\fullv{The next lemma shows that $K_0$ has the required property.}

\lem \label{l0}
Let $K_0$ be defined as above. If the DM plays $a_0$ $K_0$ times at
state $s$, then with probability $\ge 1-\delta/4N$ a new action will be
discovered if there is at least one new action at state $s$ to be
discovered. 
\elem

%nan21: should we add back this proof too?
%joe14:
\fullv{
\prf
%joe8
%For all state $s$ where the DM is unaware of at least one new action,
%assume there are $j_s$ new actions (note that the value of $j_s$ is
%unknown to the DM), we have  
Suppose that $s$ is a state where the DM is unaware of $j \ge 1$
actions.  Then
$$\begin{array}{llll}
Pr(E_{K_0,s})	&=&		\prod_{t'=1}^{K_0}(1-D(j_s,t')) \\
&\leq&	\prod_{t'=1}^{K_0}(1-D(1,t')).\\		
\end{array}$$
%joe8: 
%Since $D(1,t') \ge 0$, we show below that $1 - D(1,t') \le 
%e^{-D(1,t')}$.  Thus, we get that  
We show below that $1 - D(1,t') \le 
e^{-D(1,t')}$.  Thus, 
%joe9: shortened
\fullv{
$$\begin{array}{lll}
\Pr(E_{K_0,s}) &\le & \prod_{t'=1}^{K_0} e^{-D(1,t')}\\
	&\le &e^{-\sum_{t'=1}^{K_0} D(1,t')}.
\end{array}$$} 
%joe14: put inline to save space
%\shortv{$$\Pr(E_{K_0,s}) \le  \prod_{t'=1}^{K_0} e^{-D(1,t')}
\shortv{$\Pr(E_{K_0,s}) \le  \prod_{t'=1}^{K_0} e^{-D(1,t')}
\le e^{-\sum_{t'=1}^{K_0} D(1,t')}$.}
%joe8
%\end{array}}$$
%joe9: removed paragraph break
%
%joe8
%By assumption, we have chosen $K_0$ such that $\sum_{t=1}^{K_0}
%D(1,t)\ge \ln(4N/\delta)$, thus 
The choice of $K_0$ guarantees that $\sum_{t=1}^{K_0}
D(1,t)\ge \ln(4N/\delta)$. Thus,
%joe9: shortened
\fullv{
$$\begin{array}{lll}
%joe8: removed line break, to save space
%\Pr(E_{K_0,s})&\le &e^{-\ln(4N/\delta)}\\
%						&=& \frac{\delta}{4N}.\\
\Pr(E_{K_0,s})&\le &e^{-\ln(4N/\delta)}= \frac{\delta}{4N}.
\end{array}$$
}
\shortv{$\Pr(E_{K_0,s})\le e^{-\ln(4N/\delta)}= \frac{\delta}{4N}$.}

It remains to show that $1-D(1,t') \le e^{-D(1,t')}$. 
%nan20: put the detailed proof for 1-x\le e^{-x} into long version,
%according to the reviews 
%joe14: 
\shortv{ 
%This is true because $1-x \le e^{-x}$ for $x \ge 0$, and $D(1,t')\ge 0$.
Since $D(1,t') \ge 0$, this follows from the more general claim that 
$1-x \le e^{-x}$ for $x \ge 0$, which can be proved using standard
calculus.  We leave details to the full paper.
}
\fullv{
Since $D(1,t')\ge 0$, it suffices to show that 
$1-x \le e^{-x}$ for $x \ge 0$.
Let $g(x) = 1-x - e^{-x}$.  
We want to show that $g(x) \le 0$ for $ x \ge 0$.
An easy substitution shows that 
$g(0) = 0$.  Differentiating $g$, we get that 
$g'(x) = -1 +e^{-x}\le0$ when  $x\ge 0$.
Since $g(0) = 0$ and $g$ is nonincreasing when $x\ge 0$, 
we must have  $g(x) \le 0$ for $x \ge 0$, as desired.
}%end fullv
\eprf
}
%joe14: \end{fullv}

%nan9
%joe8*: There are many, many problems here, including the discussion of
%the theorem, the statement of the theorem (which does not clearly
%explain the role of the parameters - there is no fixed MDPU; there's
%just an algorithm), and, most important, the proof, which I simply
%cna't make sense of.  I started doing some rewriting, but this needs a
%lot more work.
%With the correct choice of $K_0$, we now show that if all the input
%estimates are correct, the $\URMAX$ algorithm does the right thing.  
%joe14
\fullv{We first show that $\URMAX(S_0, g_0, D, N, k, R_{\max},T,\epsilon,\delta,s_0)$}
\shortv{In the full paper, we show that $\URMAX(S_0, g_0, D, N, k, R_{\max},T,\epsilon,\delta,s_0)$}
 is correct provided that the parameters are
correct.  
%
%nan9*: moved from below to here and rewrote
%nan3: rewritten with more general conditions. only sufficient proofs
%are given here, but there is still a chance that it is also a necessary
%condition.  
%nan5*: rewrote the statement and the proof of the theorem
%nan8*: changed the algorithm into a general algorithm (that applies to 
% polynomial/exponential/existence algorithms)
%nan21: should we add back the theorem below?
%joe14:
\fullv{
\thm\label{t3}
%joe8*: rewrote theorem statement. There is no single underlying MDP;
%there are many
%Let $M=(S, A, S_0,  g, g_0, P, R, R^+, R^-, D)$ be an MDPU, and let
%$M^u= (S, A, P, R)$ be the underlying MDP. Let $0<\delta<1$, and
%$\epsilon>0$ be constants. The estimated values $N$, $k$, $R_{\max}$ and
%$T$ are given to the DM as additional inputs. If these estimates are
%correct, then with probability of no less than $1-\delta$ the $\URMAX$ 
%algorithm will attain an expected return of $Opt(M^u,\epsilon,
%T)-2\epsilon$ within a number of steps polynomial in $N$, $k$, $T$,
%$\frac{1}{\epsilon}$, $\frac{1}{\delta}$ $R_{\max}$ and $K_0$. 
%nan11: shall we move 'for all MDPs...' to the front, so as to avoid a
%possible confusion  
% that a single policy works for all underlying MDPs?
%joe10: reorganized and rewrote somewhat; 
%For all states $s_0 \in S_0$, for all possible underlying MDPs
%$M=(S,A,P,R)$ such that  
%$|S| \le N$, $|A| \le k$, $R(s,s',a) \le R_{\max}$ for all $s,s' \in S$
For all MDPs $M=(S,g,A,P,R)$ 
%joe10:
compatible with $S_0$, $g_0$, $N$, $k$, $R_{\max}$, and T
(i.e., $S \supseteq S_0$, $g(s)
\supseteq g_0(s)$ for all $s \in S_0$, 
$|S| \le N$, $|A| \le k$, $R(s,s',a) \le R_{\max}$ for all $s,s' \in S$
and $a \in A$, and the $\epsilon$-return mixing time of $M$ is  $\le
T$), and all states $s_0 \in S_0$, with probability at least $1-\delta$,
$\URMAX(S_0, g_0, D, N, k, R_{\max},T,\epsilon,\delta,s_0)$ 
%joe10
%generates a
running on $M$ returns a
%nan11: changed as to make it clear that the output is a policy. $\URMAX$ computes the policy
% which has the near-optimal expected payoff, but the policy does not obtain that 
% payoff immediately. However, the output policy will be used in $\URMAX$'
% to obtain  
% the expected payoff.
%policy that has an expected return at least
policy whose expected return is at least
$\Opt(M,\epsilon,T)-2\epsilon$.   Moreover, it
does so in time polynomial in $N$, $k$, $T$,
%For all states $s_0 \in S_0$, with probability at least $1-\delta$,
%$\URMAX(S_0, g_0, D, N, k, R_{\max},T,\epsilon,\delta,s_0)$ generates a
%%policy that has an expected return at least
%policy whose expected return is at least
%$\Opt(M,\epsilon,T)-2\epsilon$ for all MDPs $M=(S,A,P,R)$ such that 
%$|S| \le N$, $|A| \le k$, $R(s,s',a) \le R_{\max}$ for all $s,s' \in S$
%and $a \in A$, and $\epsilon$-return mixing time of $T$.  Moreover, it
%does so in time polynomial in $N$, $k$, $T$,
$\frac{1}{\epsilon}$, $\frac{1}{\delta}$, $R_{\max}$, and $K_0$. 
\ethm
\prf
%joe8*: I can't follow the proof as written.   I started rewriting it,
%but you'll have to fill in the details.
%First, we need to show that the expected reward is as stated. By
%assuming that all states and actions are discovered, this part of the
%proof becomes the same as in $\RMAX$ \cite{BT02}. (Related results are
%also given in \cite{AN05ICML}, \cite{KS02}, \cite{KK99}, \cite{KKL03}
%and \cite{BT02}) 
The basic structure of the proof follows lines similar to the
correctness proof of $\RMAX$ \cite{BT02}.  (Related results are
%joe11
%proved in \cite{AN05ICML,KK99,KKL03,KS02}.)  We sketch the details here.
proved in \cite{AN05ICML,KKL03,KK99,KS02}.)  We sketch the details here.

%joe8*: Nan, what I wrote may not be quite right, but it should be the
%right structure
The running time is clear from the description of the algorithm.
Let $M_r = (S_r,A_r,g_r,P_r,R_r)$ be the MDP that is finally computed by
$\URMAX$ in execution $r$ of the policy.  Thus, $S_r$ is the set of states
discovered in execution $r$, $A_r$ is the set of actions discovered, and
so on.  Although $M_r$ may not be identical to $M$, 
we will show that the set of executions $r$ where $\Opt(M_r,\epsilon,T)
\ge \Opt(M,\epsilon,T) - 2\epsilon$ has probability at least
$1-\delta$, where the probability of an execution is determined by the
%nan12
transition probabilities $P$ of the actual MDP $M$.  The key points of
the argument 
%transition probabilities of the MDP $M$.  The key points of the argument
are the following.
\begin{enumerate}
\item[(a)] With probability at least $1-\delta/4$, every action that can be
%joe10
%played in a state $s \in S_r$ is discovered (i.e., is in $g_r(s)$).
\fullv{played in a state $s \in S_r$ is discovered (i.e., is in
$g_r(s)$).}
\shortv{played in a state $s \in S_r$ is in $g_r(s)$).}
%joe8*: I think we need something like this, as do BT.
\item[(b)] If all actions that can be played in $s \in S_r$ are
%nan10*: explained 'close to'
discovered, then with probability at least $1-\delta/4$, $P_r$ is 
%joe9
%close to $P$, more specifically $|P_r(s,s',a)-P(s,s',a)|\le
close to $P$; specifically, $|P_r(s,s',a)-P(s,s',a)|\le
\frac{\epsilon}{4NTR_{\max}}$ for all  
$s,s'\in S_r$, $a\in g_r(s)$.

%discovered, then with probability at least $1-\delta/4$, $P_r$ is close
%to $P$.  [[NAN, YOU NEED TO EXPLAIN WHAT THIS MEANS.]] 
\item[(c)] $S_r$ contains all the ``significant'' states in $S$; if (a)
and (b) hold, with probability at 
least $1-\delta/4$, 
$\Opt(M_r,\epsilon,T) \ge \Opt(M,\epsilon,T) - 2\epsilon$.
%nan10: we are fine with (a) (b) and (c)
%\item[(d)] ??
\end{enumerate}

Part (a) is immediate from Lemma~\ref{l0} and the fact that $|S_r| \le N$.
%nan10
Parts (b) and (c) are similar to the arguments given by Brafman
%Parts (b), (c), and (d) are similar to the arugments given by Brafman
and Tennenholtz, so we defer the details to the full paper.
%joe8
%By making the failure rate less than $\frac{\delta}{4}$ in each case, we
%are able to obtain a total failure probability of no more than
%$\delta$. 
%joe9: Thinking about it, I'm not so sure it's ``easily''.
%The desired result now follows easily.
The desired result now follows using techniques similar in spirit to
%joe11
%those used by (Brafman and Tennenholtz, 2002); we leave the
those used by Brafman and Tennenholtz \citeyear{BT02}; we leave the
details to the full paper.
\eprf

%According to the above analysis, the algorithm requires at most $K_2$
%attemps to learn each state-action entry. Since there are at most $Nk$
%entries to learn, the algorithm spends at most $K_2Nk$ time to learn the
%entire MDP. After the underlying MDP has been learned, the algorithm
%needs to run the optimal policy for at most $z$ iterations to obtain the
%expected reward. Since $K_2$ and $z$ are polynomial in the input
%parameters, the running time for the algorithm is polynomial in $N$,
%$k$, $T$, $1/\epsilon$ and $1/\delta$. Observe that in the actual
%implementation of $\URMAX$, the only bounds that needs to be considered
%are $\tilde{K}^{a_0}$ and $K_1$. 
%\eprf
%
%joe8*:
%The above theorem shows that when the estimates on $N$, $k$, $R_{\max}$
%and $T$ are correct, $\URMAX$ works correctly. We now provide an algorithm
%$\URMAX$' which works correctly even when the DM does not know any of $N$,
%$k$, $R_{\max}$ and $T$. 
%
%Given inputs $(S_0, g_0, D, R^+, R^-)$, $\epsilon$ and $\delta$,
%$\URMAX$' proceeds as follows: 
}
%joe14: \end{fullv}
%joe14
%We now define $\URMAX(S_0,g_0,D,\epsilon,\delta,s_0)$.  The idea is to run 
We get $\URMAX(S_0,g_0,D,\epsilon,\delta,s_0)$ by running
%joe14: typo!
%$(S_0, g_0, D, N, k, R_{\max},T,\epsilon,\delta,s_0)$ using larger and
$\URMAX(S_0, g_0, D, N, k, R_{\max},T,\epsilon,\delta,s_0)$ using larger and
larger values for $N$, $k$, $R_{\max}$, and $T$.  Sooner or later the
right values are reached.  Once that happens, with high probability, the
policy produced will be optimal in all later iterations.  However, since
we do not know when that happens, we need to continue running the
algorithm.  We must thus play the optimal policy computed at each
iteration enough times to ensure that, if we have estimated $N$, $k$,
$R_{\max}$, and $T$ correctly, the average reward stays within
%nan12: typo
$2\epsilon$ of optimal while we are testing higher values of these parameters.
%$\epsilon$ of optimal while we are testing higher values of these parameters.
For example, suppose 
%joe14
%for simplicity
that the actual values of these
parameters are all 100.  Thus, with high probability, the policy
computed with these values will give an expected payoff that is within
$2\epsilon$ of optimal.  Nevertheless, the algorithm will set these
parameters to 101 and recompute the optimal policy.  While this
recomputation is going on, it may get low reward (although, eventually
it will get close to optimal reward).  We need to ensure that this
%nan12: it desn't affect the average reward in fact, according to our
%computation of $K_2+K_3$. 
%period of low rewards does not affect the average too badly.
period of low rewards does not affect the average.

%joe8: Nan, this algorithm is not correct.  You have to repeat the
%computation enough times, and you need to calculate what ``enough'' is.

\begin{tabbing}
\fullv{$\URMAX(S_0,g_0,D,\epsilon,\delta,s_0)$:\\ \\}
\shortv{$\URMAX(S_0,g_0,D,\epsilon,\delta,s_0)$:\\}
%nan20: adjusted the body of the algorithm to suit the wider column of NIPS format
%nan21: changed algorithm back to UAI format
\commentout{
\quad Set $N:=|S_0|$, $k:=|A_0|$, $R_{\max}:=1$, $T:=1$\\
\quad {\bf Repeat forever}\\
\quad \quad \= Run $\URMAX((S_0, g_0, D,N,k,R_{\max},T,\epsilon,\delta,s_0)$\\
\quad \quad \= {\bf if} no inconsistency is discovered {\bf then} run the policy computed by\\ 
\quad \quad \quad $\URMAX((S_0, g_0, 
%joe15
%D,N,k,R_{\max},T,\epsilon,\delta,s_0)$ for $K_2 + K_3$ steps, where\\
%\quad \quad \quad \quad $K_2 =
D,N,k,R_{\max},T,\epsilon,\delta,s_0)$ for $K_2 + K_3$ steps,\\
\quad \quad \quad \quad where $K_2 =
2(Nk\max(K_1(T+1),K_0))^{\frac{3}{2}}R_{\max}/\epsilon$\\
\quad \quad \quad \quad $K_3 =
(2R_{\max}
+1)\max((\frac{2R_{\max}}{\epsilon})^3,8\ln(\frac{4}{\delta})^3)/\epsilon$\\ 
\quad \quad \=$N := N+1$; $k := k+1$, $R_{\max} := R_{\max}+1$, $T := T+1$.
}
%nan10: initialization  added
%joe9: used := insead of = 
Set $N:=|S_0|$, $k:=|A_0|$, $R_{\max}:=1$, $T:=1$\\
%joe8: I think it's clearly if we insert all the parameters; rewrote algorithm
%\bf{$\URMAX$'}\\ \\
%Repeat\\
%\quad \= $\URMAX((S_0, g_0, D, R^+, R^-),N,k,R_{\max},T,\epsilon,\delta)$\\
%\quad \=Set $N+1$, $k=k+1$, $R_{\max}=R_{\max}+1$, $T=T+1$.
{\bf Repeat forever}\\
\quad \= Run $\URMAX((S_0, g_0, D,N,k,R_{\max},T,\epsilon,\delta,s_0)$\\
\quad \= {\bf if} no inconsistency is discovered\\
\quad \quad {\bf then} run the policy computed by\\ 
\quad \quad $\URMAX((S_0, g_0, 
%joe8*: Nan, you need to compute the exact number of steps
%nan10*: number of steps added
%joe9: changed and simplified notation; First, we haven't
%defined K_2 yet, so it's funny to use K_3; I wrote K_2 + K_3 instead.
%Second, in a_3, there no reason to have the numerator as R_max -
%2\epsilon; it's surely OK to have just R_max.  Finally, it seems that
%the only difference between K_1' and K_1 is that T+1 is used instead of
%T.  My guess
%nan11: you are exactly right. K_1' is in fact just K_1(T+1).
%D,N,k,R_{\max},T,\epsilon,\delta,s_0)$ for $K_3$ steps\\
D,N,k,R_{\max},T,\epsilon,\delta,s_0)$ for\\ 
%joe16: undid change; I think it looks better with ``where on earlier line
%nan22: changed according to Joe's comment
\quad \quad  $K_2 + K_3$ steps, where\\
%\quad \quad  $K_2 + K_3$ steps,\\
\quad \quad \quad where $K_2 = 
%D,N,k,R_{\max},T,\epsilon,\delta,s_0)$ for ** steps\\
%nan10: variable definitions
%joe9*
%\quad \quad where $K_3=a_2+a_3$ such that:\\
%\quad \quad $\ \
%$$a_2=\max((\frac{2R_{\max}}{\epsilon})^3,8\ln(\frac{4}{\delta})^3)+1$\\  
%\quad \quad $a_3=\frac{2(a_1+a_2)(R_{\max}-2\epsilon)}{\epsilon}$\\
%\quad \quad $\ \ $ $a_1= (Nk\max(K_1',K_0))^{\frac{3}{2}}$\\
%\quad \quad 
%$K_1'=\max(\lceil\frac{4N(T+1)R_{\max}}{\epsilon}\rceil^3,$
%$\lceil-8\ln^3(\frac{\delta }{8Nk})\rceil)+1$\\ 
%\max((\frac{2R_{\max}}{\epsilon})^3,8\ln(\frac{4}{\delta})^3)+1$\\   
%\quad \quad $a_3=\frac{2(a_1+a_2)(R_{\max}-2\epsilon)}{\epsilon}$\\
%\quad \quad $\ \ $ $a_1= (Nk\max(K_1',K_0))^{\frac{3}{2}}$\\
%\quad \quad $\ \ $
%$K_1'=\max(\lceil\frac{4N(T+1)R_{\max}}{\epsilon}\rceil^3,$
%$\lceil-8\ln^3(\frac{\delta }{8Nk})\rceil)+1$\\ 
%end nan10: variable definitions
%joe9*: notice it's K_1(T+1); I think this is just your K_1'; I've also
%added the 2/\epsilon
%\quad \quad \quad $K_2 =
2(Nk\max(K_1(T+1),K_0))^{\frac{3}{2}}R_{\max}/\epsilon$\\
\quad \quad \quad $K_3 =
%joe9*: the +1 at the end is surely unnecessary.  I'm also not sure we
%need 2R_{\max} + 1; I suspect that 2R_{\max} should be enough.
(2R_{\max}
+1)\max((\frac{2R_{\max}}{\epsilon})^3,8\ln(\frac{4}{\delta})^3)/\epsilon$\\ 
\quad \=$N := N+1$; $k := k+1$, $R_{\max} := R_{\max}+1$, $T := T+1$.
\end{tabbing}
%nan21: end algorithm
%joe8*: we need to state a formal theorem
%In $\URMAX$', we repeatedly carry out $\URMAX$ with incrementally increasing
%values of $N$, $k$, $R_{\max}$ and $T$. We have shown in Theorem
%\ref{t3} that if the estimates of these values are correct (i.e. are
%upper bounds of the true values), $\URMAX$ works correctly. Thus, sooner
%or later, all variables will become upper bounds of the true values, and
%from then on, $\URMAX$' will continue to work correctly. 
%
%It is easy to see that $\URMAX$' requires a polynomial (in fact linear)
%number of iterations in $N$, $k$, $R_{\max}$ and $T$ to get the correct
%value of estimates. Since $\URMAX$ is a polynomial algorithm, thus,
%$\URMAX$' computes a near-optimal policy in time polynomial in $N$, $k$,
%$T$, $\frac{1}{\epsilon}$, $\frac{1}{\delta}$, $R_{\max}$ and $K_0$. 

%nan9: cut
%nan2: arguments for removing the bounds on number of actions and states
%We can remove the bound on the number of actions by doing the
%following. At each state, it requires $a_0$ to cotinuously fail for at
%most $K_0$ times before the algorithm stops to look for new actions at
%that state.  However, when  $K_0$ does not depend on $k$ (as is the
%case in Theorem \ref{t1}, \ref{t4} and \ref{t5}), we can determine the
%actual value of $k$ after playing $a_0$ for a sufficient number of
%times (in fact, at most $kK_0$ times since we play $a_0$ for at most
%$K_0$ times for discovering each new action). After $k$ is determined,
%We then execute the algorithm as normal.  

The following theorem shows that $\URMAX(S_0,g_0,D,\epsilon,\delta,s_0)$
%nan10: added
is correct.  
%joe9:
(The proof, which is deferred to the full paper, explains the choice of
$K_2$ and $K_3$.)
\thm\label{t4} For all MDPs $M = (S,A,g,P,R)$ compatible with $S_0$ and
$g_0$, if  
the $\epsilon$-return mixing time of $M$ is $T_M$, then for all states
$s_0 \in S_0$, with probability at least $1-\delta$, for all states $s_0
\in S_0$, $\URMAX(S_0,g_0,D,\epsilon,\delta,s_0)$ computes a policy
$\pi_{\epsilon,\delta,T_M,s_0}$ such that, for a time
$t_{M,\epsilon,\delta}$ that is  
polynomial in $|S|$, $|A|$, $T_M$, $1/\epsilon$, and $K_0$, and all $t \ge
t_{M,\epsilon,\delta}$, we have $U_M(s_0,\pi,t) \ge \Opt(M,\epsilon,T_M) -
%nan12: should be $2\epsilon$
2\epsilon$.  
%\epsilon$.  
\ethm
%The bound on the number of states can be removed too. Initialize $N$ to
%some value $\geq |S_0|$, and execute the algorithm as normal. Then
%incrementally increase the value of $N$. The algorithm achieves a
%near-optimal reward when $N$ becomes greater than the real number of
%states. After that, it continues to achieve a near-optimal reward.  
%end nan2

Thus, if $\sum_{t=1}^\infty D(1,t) = \infty$, the DM can learn to play
near-optimally.   We now get running time estimates that essentially
match the lower bounds of Theorem~\ref{t2}.

%joe14: upgraded to proposition
%\lem If $\sum_{t=1}^T D(1,t) \ge f(T)$, where $f : [1,\infty]
\pro If $\sum_{t=1}^T D(1,t) \ge f(T)$, where $f : [1,\infty]
\rightarrow \IR$ is an increasing function whose co-domain includes
$(0,\infty]$, then $K_0 \le f^{-1}(\ln(4N/\delta))$, and the running
time of $\URMAX$ is polynomial in $f^{-1}(\ln(4N/\delta))$.
%joe14
%\elem
\epro
%joe14
\fullv{
\prf Immediate from Theorem~\ref{t4} and the definition of $K_0$. \eprf

Recall from Theorem~\ref{t3} that if $\sum_{t=1}^T D(1,t) \le f(T)$, the
no algorithm that learns near-optimally can run in time less than
%nan10: changed fist c to c', since previously they are both called
%'c'. also fixed typo in the expression.  
$f^{-1}(c'\ln(1/\delta))$ (where $c' = c/\ln(1/(1-c))$), so we have proved an
%$f^{-1}(c\ln(1/\delta))$ (where $c = \ln(1/(1-c))$), so we have proved an
upper bound that essentially matches the lower bound of Theorem~\ref{t3}.
}
%joe14: \end{fullv}

%joe12: added label
%\cor If $\sum_{t=1}^T D(1,t) \ge m_1 \ln(T) + m_2$ (resp.,
\cor\label{cor:poly} If $\sum_{t=1}^T D(1,t) \ge m_1 \ln(T) + m_2$ (resp.,
%nan10: added constraints on m_1 and m_2
$\sum_{t=1}^T D(1,t) \ge m_1 \ln (\ln(T) + 1) + m_2$) for some constants $m_1>0$ and $m_2$, then 
%$\sum_{t=1}^T D(1,t) \ge m_1 \ln (\ln(T) + 1) + m_2$), then 
the DM can learn to play near-optimally in polynomial time (resp.,
exponential time).
\ecor
%joe9: left to full paper
\fullv{
\prf If $f(T) = m_1 \ln(T) + m_2$, then as we have observed, $f^{-1}(t)
= e^{(t-m_2)/m_1}$, so $f^{-1}(\ln(4N/\delta) =
%nan10: typo, and added a space for breaking into two lines 
e^{-m_2/m_1}(4N)^{1/m_1}\cdot$ $(1/\delta)^{1/m_1}$.  Thus,
%e^{m_1/m_2}(4N)^{1/m_1}(1/\delta)^{1/m}$.  Thus,
%nan10:  
$f^{-1}(\ln(4N/\delta)$ has the form $a(1/\delta)^{1/m_1}$, and is
%$f^{-1}(\ln(4N/\delta)$ has the form $a(1\delta)^{1/m}$, and is
polynomial in $1/\delta$.  The result now follows from
Theorem~\ref{t3}.  The argument is similar if $\sum_{t=1}^T D(1,t) \ge
m_1 \ln (\ln(T) + 1) + m_2$; we leave details to the reader.
\eprf
}

%joe12: tried to deal with it earlier.  Let me know what you think.
\commentout{
% ASHUTOSH: One of the robotics example in the beginning described
% the impossibility case--- when a designer would rather have human
% designers design the actions.  Here we need to tie it back to the
% polynomial time case by writing a small paragraph.  E.g.:
This setting applies when the discovery probability is high
that it is possible for the robot to discover actions by itself. 
For example, in the case of a helicopter trying to discover actions,
the space of actions is four dimensional (which affects the 6 dof
state of the helicopter); therefore our results show that it is 
possible to learn it automatically. (Relate as to exactly how?)
Optional: Indeed, in prior work ...
}

%nan13: conclusion added back, to give the readers a better feeling of
%what we've 
% done, how it could be applied to real applications, it also provides the 
% direction of future works
%joe11: To get it down to 8 pages, I cut this for now
%joe15: added back in
%\fullv{
%\section{Conclusion}\label{sec:conc}
\section{CONCLUSION}\label{sec:conc}
%joe15
%We have defined an extension of MDPs that we call MDPUs to deal with the
We have defined an extension of MDPs that we call MDPUs, Markov Decision
Processes with Unawareness, to deal with the
possibility that a DM may not be aware of all possible actions. We provided a 
complete characterization of when a DM can learn to play near-optimally
in an MDPU, and have provided an algorithm that learns to play
near-optimally when it is possible to do so, as efficiently as possible.
%joe15: added (this sentence was commented out earlier in the paper;
%this is the right place for it)
Our methods and results
thus provide guiding principles for designing complex systems.

We believe that MDPUs should be widely applicable.  We hope to apply the
insights we have gained from this theoretical analysis to using MDPUs in
%joe15
%practice. For example, to enable a robotic car to learn new driving
%skills that are initially unaware to the robot.
practice, for example, to enable a robotic car to learn new driving skills.
%joe16
%Our results show that there will be time when an agent cannot hope to
%is what the agent should do, if he has a limited budget.  Work on
Our results show that there will be situations when an agent cannot hope to
learn to play near-optimally.  In that case, an obvious question to ask
is what the agent should do.  Work on
budgeted learning has been done in the MDP setting 
%joe15
%\cite{GKN09}, \cite{MLG04}, \cite{GM07};  
\cite{GKN09,GM07,MLG04};  
we would like to extend this to MDPUs.
%}
%joe8: \end{fullv}

%joe15: added funding acknowledgements
\paragraph{Acknowledgments:} The work of Halpern and Rong was supported
in part by NSF grants IIS-0534064, IIS-0812045, and
IIS-0911036, and by AFOSR grants 
FA9550-08-1-0438 and FA9550-09-1-0266, and ARO grant W911NF-09-1-0281.

%nan23
\fullv{

}

\end{document}